\documentclass{CVM}

\usepackage[numbers]{natbib}
\usepackage{graphicx}
\usepackage{geometry}
\usepackage{amsmath}

\usepackage{booktabs,makecell}
\usepackage{bbding}
\usepackage{pifont}
\usepackage{wasysym}
\usepackage{amssymb}
\usepackage{booktabs}
\usepackage{caption}
\usepackage{arydshln}

\usepackage{mwe}
\usepackage[normalem]{ulem}
\usepackage{enumitem}

\usepackage{url}
\usepackage{xcolor}
\usepackage{hyperref}
\usepackage{footnote}

\newcommand{\ie}{{\it i.e.}}
\newcommand{\eg}{{\it e.g.}}

\newcommand{\cmark}{\ding{51}}%
\newcommand{\xmark}{\ding{55}}%
\definecolor{mypink1}{rgb}{0.858, 0.188, 0.478}
\definecolor{mygray}{gray}{0.6}

\usepackage{algorithm,algpseudocode}
\algnewcommand{\Inputs}[1]{%
\Statex \textbf{Inputs:}
  \Statex \hspace*{\algorithmicindent}\parbox[t]{.8\linewidth}{\raggedright #1}
}

\algnewcommand{\Outputs}[1]{%
\Statex\textbf{Outputs:}
  \Statex \hspace*{\algorithmicindent}\parbox[t]{.8\linewidth}{\raggedright #1}
}

\algnewcommand{\algorithmicforeach}{\textbf{for each}}
\algdef{SE}[FOR]{ForEach}{EndForEach}[1]
  {\algorithmicforeach\ #1\ \algorithmicdo}% \ForEach{#1}
  {\algorithmicend\ \algorithmicforeach}% \EndForEach

\CVMsetup{
type      = {ResearchArticle},
doi       = {s41095-0xx-xxxx-x},
title     = {A Unified Multi-view Multi-person Tracking Framework},
author    = {Fan Yang$^{1}$\cor{fan.yang@fujitsu.com}, Shigeyuki Odashima$^{1}$, Sosuke Yamao$^{1}$,  Hiroaki Fujimoto$^{1}$,  Shoichi Masui$^{1}$,  and Shan Jiang$^{1}$},
runauthor = {F.Y., S.O., S.Y., H.F., S.M., S.J.},
abstract  = {
  Although there is a significant development in 3D Multi-view Multi-person Tracking (3D MM-Tracking), current 3D MM-Tracking frameworks are designed separately for footprint and pose tracking. Specifically, frameworks designed for footprint tracking cannot be utilized in 3D pose tracking, because they directly obtain 3D positions on the ground plane with a homography projection, which is inapplicable to 3D poses above the ground. In contrast, frameworks designed for pose tracking generally isolate multi-view and multi-frame associations and may not be robust to footprint tracking, since footprint tracking utilizes fewer key points than pose tracking, which weakens multi-view association cues in a single frame. This study presents a Unified Multi-view Multi-person Tracking framework to bridge the gap between footprint tracking and pose tracking. Without additional modifications, the framework can adopt monocular 2D bounding boxes and 2D poses as the input to produce robust 3D trajectories for multiple persons. Importantly, multi-frame and multi-view information are jointly employed to improve the performance of association and triangulation. The effectiveness of our framework is verified by accomplishing state-of-the-art performance on the \textit{Campus} and \textit{Shelf} datasets for 3D pose tracking, and by comparable results on the \textit{WILDTRACK} and \textit{MMPTRACK} datasets for 3D footprint tracking.
},
keywords  = {Multi-camera Multi-person Tracking, 3D Trajectory, Triangulation with Outlier Rejection, Spatiotemporal Clustering },
copyright = {The Author(s)},
}

\begin{document}

\maketitle

    \begin{figure}[b] \vskip -4mm
    \small\renewcommand\arraystretch{1.3}
        \begin{tabular}{p{80.5mm}} \toprule\\ \end{tabular}
        \vskip -4.5mm \noindent \setlength{\tabcolsep}{1pt}
        \begin{tabular}{p{3.5mm}p{80mm}}
    $1\quad $ & Fujitsu Research, Japan.\\ 
&\hspace{-5mm} Manuscript received: 2022-09-29; accepted: xxxx \vspace{-2mm}
    \end{tabular} \vspace {-3mm}
    \end{figure}

\section{Introduction}
\begin{figure}[!h]
  \centering
  \includegraphics[width=\linewidth]{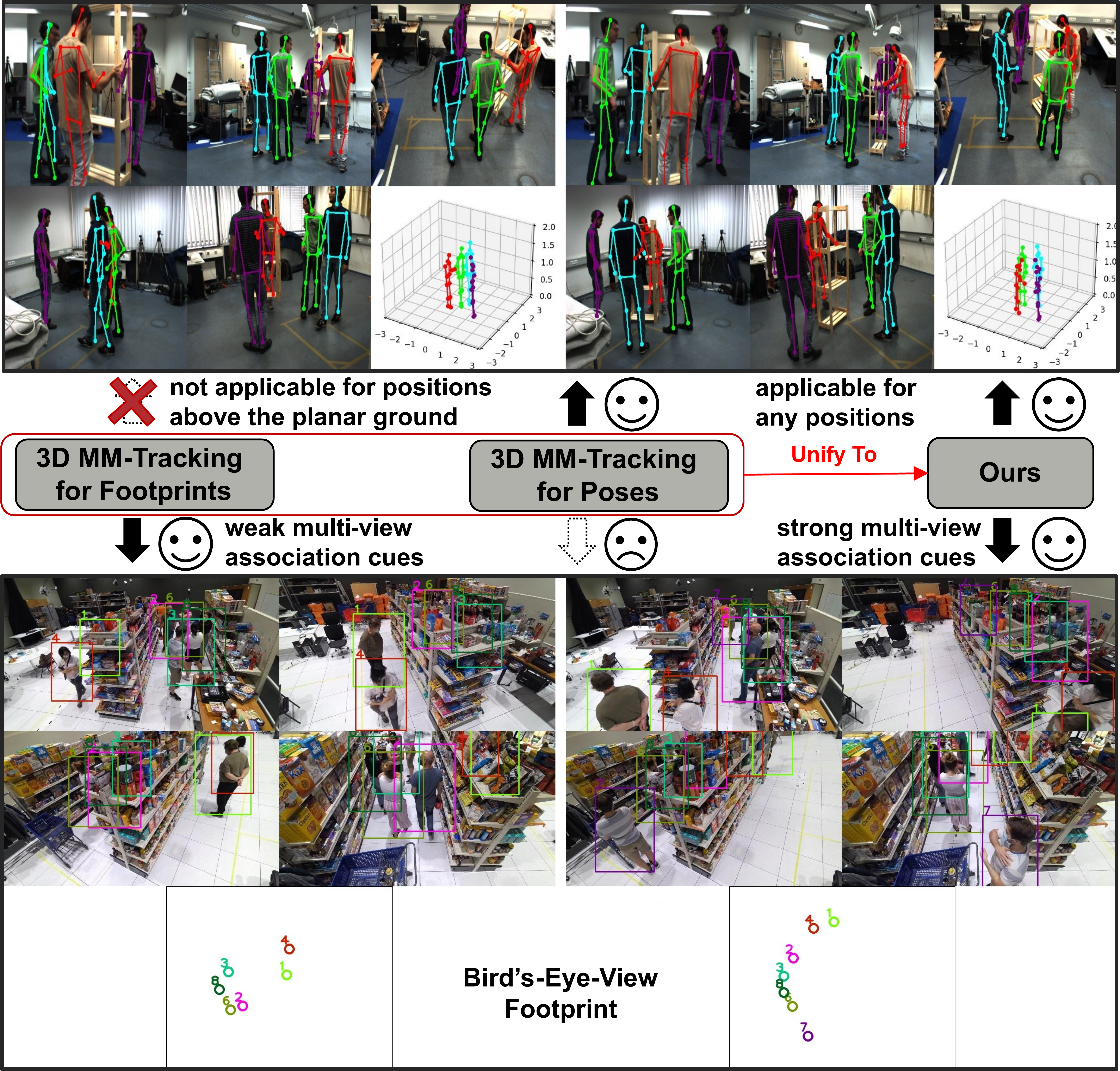}
  \captionsetup{font=small}
  \caption{\textbf{Illustration of our proposed framework.} Monocular 2D bounding boxes and 2D poses can be applied to yield 3D trajectories for multiple persons.}
  \label{fig:demo}
\end{figure}

A wide range of real-life applications (\eg, human-computer interaction) demands 3D person trajectories (\ie, footprint or pose trajectories). To obtain high-quality 3D person trajectories, multi-view camera systems have been developed to overcome challenges (\eg, occlusions) encountered in monocular systems. By leveraging multi-view information, 3D \textbf{M}ulti-view \textbf{M}ulti-Person Tracking (3D MM-Tracking) can recover partial 3D information that is not observed in a monocular view. Under the popular tracking-by-detection paradigm, 3D MM-Tracking can be approached by first obtaining camera parameters and monocular 2D features (\eg, bounding boxes or poses), and then associating multi-view 2D features to generate 3D tracklets. In this work, we focus on how to associate 2D features and generate 3D tracklets.

The cameras are synchronized and work together to identify and track people as they move around in the environment.
Depending on the application, the desired 3D person trajectory could be a 3D footprint generated from multi-view 2D bounding boxes, or, 3D pose sequences obtained from multi-view 2D poses. Although numerous works~\cite{black2006multi,sternig2011multi,he2020multi,chen2020multi,chen2020cross,ohashi20vmocap,dong2021fast,zhang2022voxeltrack} have contributed to 3D MM-Tracking, we realized that existing frameworks are designed separately---for 3D footprint tracking using 2D boxes, or, for 3D pose tracking using 2D poses.  Consequently, while those tracking methods have their limitations, their merits have also not been shared. 

On one hand, frameworks designed for tracking footprints~\cite{sternig2011multi,wen2017multi,kohl2020mta,he2020multi} appear to weakly connect multi-frame and multi-view associations to improve tracking robustness. However, these methods generally assume that individuals are on the flat ground and their footprints can be directly computed from a monocular 2D bounding box with homographic projection, which is inapplicable to 3D poses above the ground and leads to limited applicability. On the other hand, frameworks designed for tracking poses~\cite{chen2020multi,chen2020cross,ohashi20vmocap,dong2021fast} generally isolate multi-view and multi-frame associations. Because of the 2D-to-3D ambiguity and complexity of human articulation, it could encounter difficulties in associating multi-view 2D features correctly in a single frame. Moreover, performing multi-view association on 2D bounding boxes with the aforementioned methods might be more difficult than that on 2D poses, because 2D bounding boxes have fewer key points than 2D poses, which weakens multi-view association indices in a single frame~\cite{canton2005towards}.

To integrate the strengths and mitigate the limitations of the aforementioned methods, we proposed a Unified Multi-view Multi-person Tracking framework (\textit{c.f.} Figure~\ref{fig:demo}). \textit{Notwithstanding, creating a unified framework is not simply a matter of combining existing methods, but we bring new insights to enhance the robustness and efficiency of our framework}.

\tabcolsep=4pt
\begin{table}[!h]
\centering
\caption{Notation Table.}
\label{tab:notations}
\resizebox{\linewidth}{!}{
\begin{tabular}{cl}
\toprule
\textbf{Symbol} & \textbf{Description} \\ \midrule
\multirow{3}{*}{$\boldsymbol{x}_{t,i} \in \mathbb{R}^{2}$} &  2D position with index $i$ at frame $t$. Its camera ID \\ & is $c_{i}$   and its 2D coordinate is ($x_{t,i},y_{t,i}$). \\ &Its horizontal and vertical scales are $w_{t,i}$ and $h_{t,i}$ \\ \midrule
$\boldsymbol{l}_{c_{j}}(\boldsymbol{x}_{t,i})$ & Epipolar line of $\boldsymbol{x}_{t,i}$ in camera $c_{j}$ \\ \midrule
$d^{cross} (\boldsymbol{x}_{t,i}, \boldsymbol{x}_{t,j})$ & Consistency distance between cross-view 2D  positions \\ \midrule
\multirow{5}{*}{$\mathcal{T}_{k,i}$} & 2D tracklet with index $i$, it is cropped by the \\&sliding window anchored at keyframe $k$.  Its active \\ & frames form a set $\Psi_{i}$ and its camera ID is $c_{i}$ \\ &If it is assigned to cluster $p$ in multi-view \\ &association, it can be further denoted as $\mathcal{T}_{k,p,i}$ \\ \midrule
$\nu $ & Window size of sliding window\\ \midrule
$\delta $ & Step size of sliding window\\ \midrule
\multirow{2}{*}{$S(\mathcal{T}_{k,i}, \mathcal{T}_{k,j})$} & A set of distance between $\mathcal{T}_{k,i}$ and $\mathcal{T}_{k,j}$ within \\ 
&the sliding window anchored at keyframe $k$\\ \midrule
$\Omega_{k}$ & The set of clusters at keyframe $k$ \\  \midrule
\multirow{3}{*}{$\omega_{k,i}$} & Cluster at keyframe $k$ with index $i$.\\ & multi-view 2D tracklets belong to the identical \\ & person will be assigned to the cluster. \\  \midrule
\multirow{3}{*}{$\lambda$} &  Threshold of associating multi-view 2D tracklets. It is \\ & determined by the variance of our normalized  \\ & epipolar distance. \\  \midrule
$\bold{P}_{c_i}$ & Perspective projection matrix for camera $c_i$  \\ \midrule
\multirow{2}{*}{$\bold{H}_{c_i}$} & Homography projection matrix from the camera $c_i$ \\ & to the 2D ground plane  \\ \midrule
$\bold{F}_{c_i, c_j}$ & Fundamental matrix from camera $c_i$ to $c_j$ \\ \midrule
\multirow{2}{*}{$\boldsymbol{X}_{t,m} \in \mathbb{R}^{3}$} & 3D position with index $m$ at frame $t$. \\ &Its 3D coordinate is ($X_{t,m}, Y_{t,m}, Z_{t,m}$)\\ \midrule
$\varphi$ & 1D interpolation window size  \\\midrule
\multirow{3}{*}{$\mathcal{U}_{k,i}$} & 3D tracklet with index $i$, it is cropped by the \\&sliding window anchored at keyframe $k$.  Its active \\ & frames form a set $\Psi_{i}$ \\  \midrule
\multirow{3}{*}{$\kappa $} & Cutting threshold for Multi-view Multi-view \\ & Triangulation, the default value is 0.2 meters.\\ & It represents the tolerance of 3D bias \\\midrule
$\Pi $ & A set of 3D positions at one frame before outlier rejection. \\  \midrule
$\mathbf{D}$ & Distance matrix between 3D tracklets. \\  \midrule
$\mathbf{M}$ & \makecell[l]{Assignment matrix, which is a Boolean matrix.\\ When row $i$ is assigned to column $j$, $\mathbf{M}_{i,j}=1$.}
\\\bottomrule
\end{tabular}}
\end{table}

We suppose that multi-view and multi-frame associations are highly correlated: multi-view geometric constraints can exclude false detection and improve multi-frame association in a monocular view, whereas multi-frame association in each view can compensate for the effect of noise and outliers that hamper multi-view association. Therefore, we attempt to jointly utilize multi-frame and multi-view information in our framework. In detail, we traverse the entire video with sliding windows to perform online processing. In each sliding window, we first connect 2D positions to 2D tracklets, and then compute the cross-view consistency between multi-view 2D tracklets using our normalized epipolar distance, which is irrelevant to the projection variance. Using the consistency distance, multi-view 2D tracklets are associated to clusters using our \textbf{P}ropagable \textbf{D}istance-based \textbf{N}on-parametric \textbf{C}lustering (PDNC), which can propagate a calculable distance to compensate for an incalculable distance under spatiotemporal constraints. Subsequently, we obtain the 3D positions by applying our \textbf{C}ollaborative \textbf{M}ulti-frame \textbf{M}ulti-view \textbf{T}riangulation (CMMT), which consolidates multi-frame multi-view information to calculate 3D positions and reject outliers. Finally, we link 3D tracklets between sliding windows. Regardless of the use of 2D poses or 2D bounding boxes, our proposal can generate high-quality 3D person trajectories.

For better readability, we summarize the notations used throughout this study in Table~\ref{tab:notations}.

\begin{figure*}[!h]
  \centering
  \includegraphics[width=0.82\textwidth]{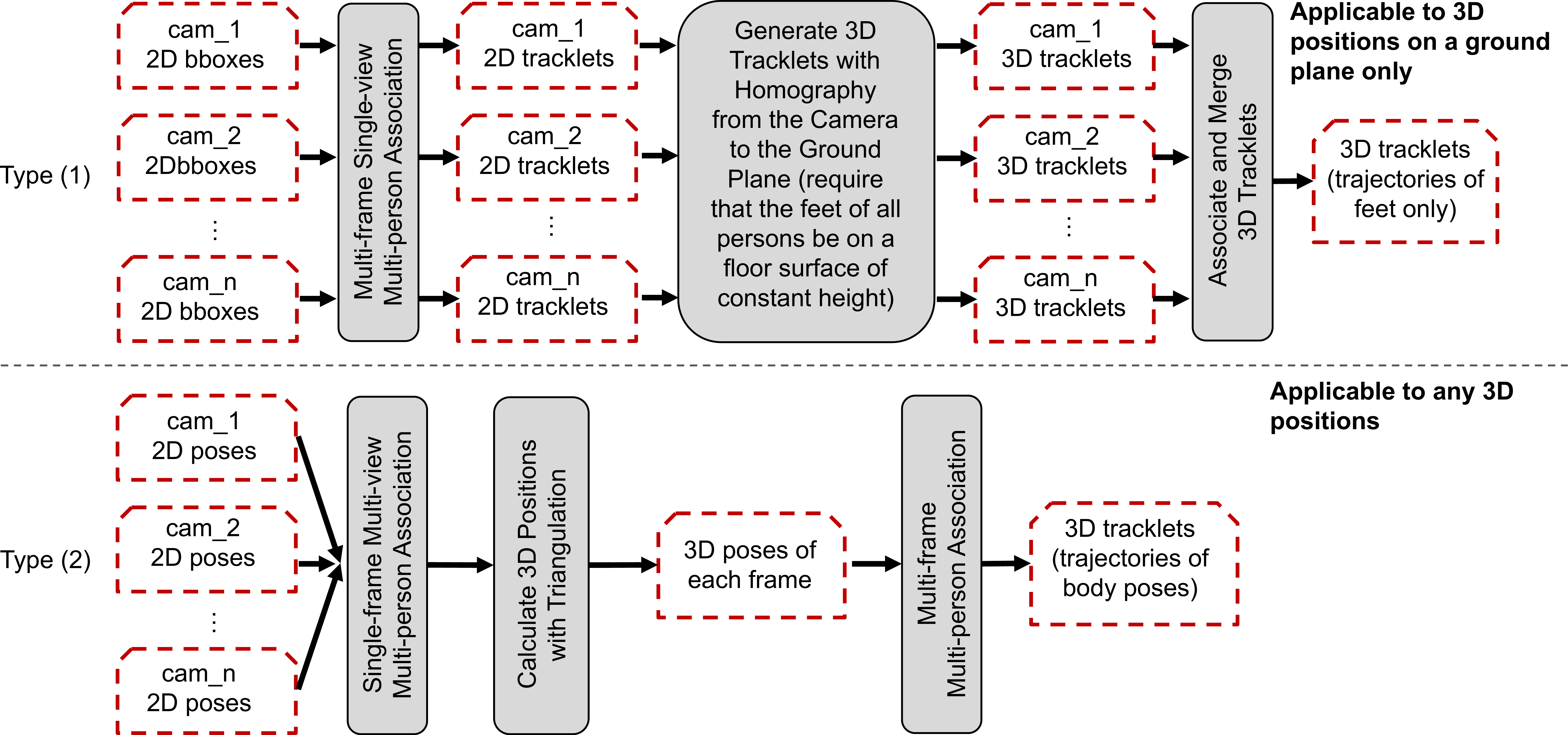}
  \captionsetup{font=small}
  \caption{\textbf{Two types of 3D MM-Tracking architectures that are related to our work.} Type (1)~\cite{sternig2011multi,wen2017multi,kohl2020mta, he2020multi} generates 2D tracklets for each view and the corresponding 3D positions are directly obtained with homography projection, which requires 3D positions on the ground plane. Then, it merges 3D tracklets projected from each view in an \textit{offline} manner. Type (2)~\cite{chen2020multi,chen2020cross,ohashi20vmocap,dong2021fast} first applies single-frame multi-view association and triangulation to obtain 3D positions, and then links 3D positions into 3D tracklets in an \textit{online} manner.}
  \label{fig:related_works}
\end{figure*}

In summary, we present the following contributions:
\begin{itemize}[noitemsep]
  \item We propose a Unified 3D MM-Tracking framework that integrates the strengths and mitigates the limitations of pose and footprint tracking. The framework also offers real-time performance on realistically sized problems.
  \item We introduce a normalized epipolar distance that, for the first time, makes cross-view consistency independent of the projected 3D-to-2D variance.
  \item We specifically design a PDNC to collaboratively associate multi-frame multi-view 2D tracklets. When cross-camera 2D tracklets do not have overlapping frames, their distance cannot be directly calculated; therefore, conventional clustering methods can not be applied to associate them. Our PDNC tackled this issue, and as a general clustering method, PDNC can also be applied to other research fields. 
  \item We present a novel yet simple CMMT, which can aggregate multi-frame multi-view information to calculate 3D positions and reject outliers. Our results prove that it is more robust than the conventional methods compared in our experiments.
  \item We propose a robust online track-to-track association to enable online processing of our framework and reduce the effect of track ID switches in tracklets of each monocular view.
  \item Our framework achieves state-of-the-art performance on the \textit{Shelf} and \textit{Campus} datasets~\cite{fleuret2007multicamera,BelagiannisAASN16}; meanwhile, it obtains comparable results on the \textit{WILDTRACK} dataset and the ICCV 2021 Multi-camera Multiple People Tracking Benchmark~\footnote{\url{https://competitions.codalab.org/competitions/33729}}. The effectiveness of our framework has been verified.
\end{itemize}

%------------------------------------------------------------------------
\section{Related Works}\label{sec:related_works}

  \subsection{2D Single-view Multi-object Tracking}

  2D Single-view Multi-object Tracking (MOT) is the basis for 3D MM-Tracking. Although it seems less complicated than 3D MM tracking, 2D single-view MOT is still challenging due to a large number of objects that need to be tracked, the occlusions between objects, and the changing appearance of objects over time. 

  In general, appearance and geometric consistency are two important assumptions used for MOT. The previous appearance of an identical object should be similar to its current appearance (\ie, appearance consistency), and its previous location and shape added to its estimated motion should be approximate to its current location and shape (\ie, geometric consistency). While appearance-based MOT methods~\cite{yang2022tackling, zeng2022motr, zhou2022global} have achieved promising performance, recent appearance-free MOT solutions~\cite{du2022strongsort, CBIOU_2023_WACV} prove that only using the geometric features can also provide robust tracking results on multiple difficult MOT datasets~\cite{giancola2022soccernet}. In this work, we recommend using the appearance-free approach to achieve fast online processing, but we also illustrate that adding an appearance feature can improve the tracking performance on some datasets.

  \subsection{3D MM-Tracking}

  We list two types of 3D MM-Tracking architectures related to our work in Figure~\ref{fig:related_works}.
  We mainly focus on how to associate multi-view 2D features and generate 3D tracklets. Therefore, we suppose that multi-view 2D observations (\eg, 2D bounding boxes or poses) are given and will not be discussed.

  \begin{figure}[!h]
    \centering
    \includegraphics[width=\linewidth]{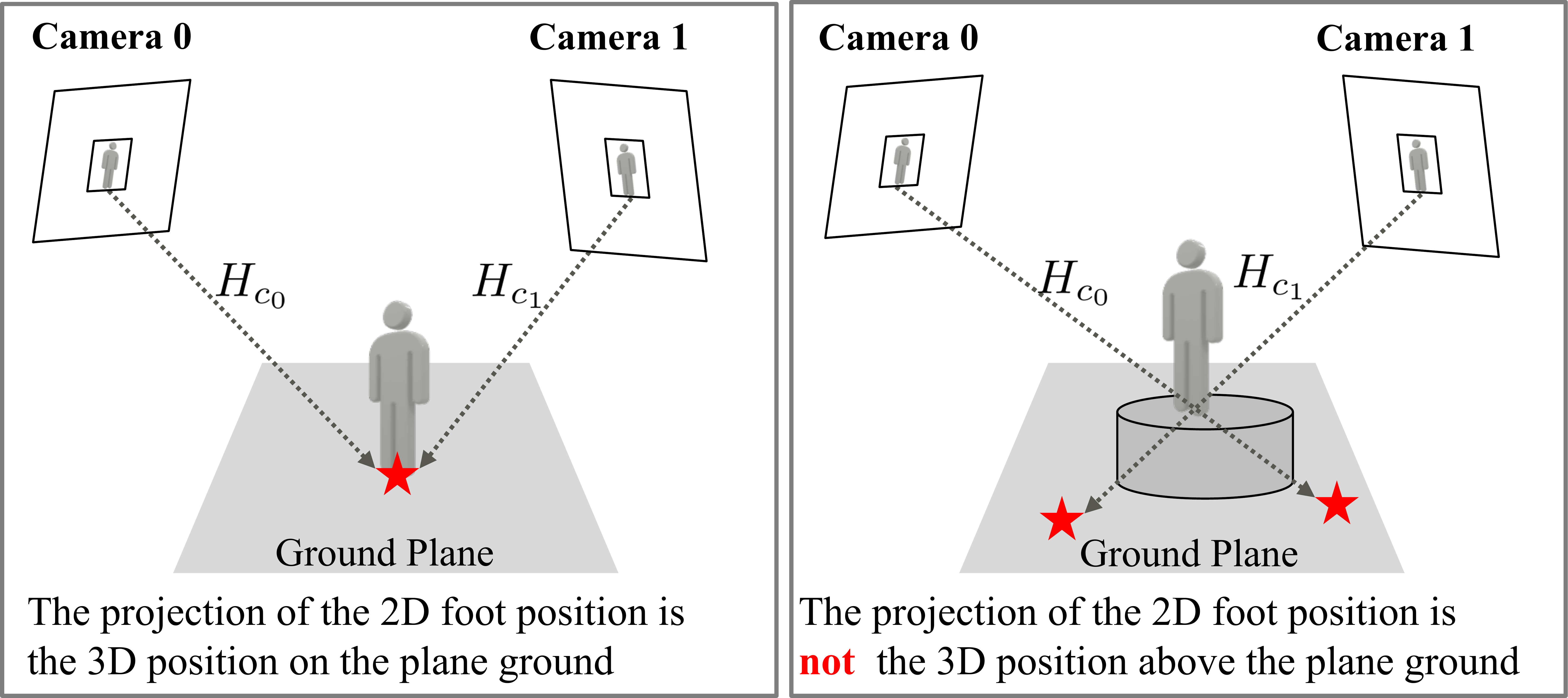}
    \captionsetup{font=small}
    \caption{\textbf{Limitation of generating the 3D position with the homography projection.} $\bold{H}_{c_i}$ is the homography projection matrix from the camera $c_i$ to the 2D ground plane. The 3D position above the ground plane cannot be obtained with the homography projection to the 2D ground plane, which was applied in previous works~\cite{sternig2011multi,wen2017multi,kohl2020mta,he2020multi}.}
    \label{fig:compare_3d_pos}
  \end{figure}

\tabcolsep=4pt
\begin{table}[!h]
\centering
\caption{\textbf{Comparison of Related Works and Our Proposal.} Where ``Bboxes'' represents 2D bounding boxes.}
\label{tab:related_works}
\resizebox{\linewidth}{!}{
  \begin{tabular}{l|c|c|c}
  \toprule
  Method & Inputs & \makecell[c]{Multi-frame Association \\\& Multi-view Association} & \makecell[c]{Target 3D \\ Positions}\\
  \midrule
  Type (1)  & Bboxes & Weakly Connected &  On The Ground  \\   \midrule
  Type (2)  & Poses & Isolated &  \makecell[c]{On and Above \\ The Ground}\\   \midrule
  Ours  & Bboxes \& Poses &  Strongly Connected & \makecell[c]{On and Above \\ The Ground}
  \\\bottomrule
  \end{tabular}}
  \end{table}

  Type (1) related works~\cite{sternig2011multi,wen2017multi,kohl2020mta,he2020multi} focused on obtaining 3D footprints. They often require all persons to stay on a planar ground with zero height. Without using multi-view stereo pairs, 3D footprints are directly generated by referring to the bottom point of the 2D bounding box in a monocular view (Figure~\ref{fig:compare_3d_pos}). In the final step, 3D tracklets mapped from different views are clustered and merged in an offline manner. \textit{However, such an offline solution could be problematic: if identity switches are encountered in a pre-obtained 2D tracklet, then such a 2D tracklet cannot be matched with others; therefore, it will be discarded entirely. Our method uses the sliding window to handle this problem. In each sliding window, the effects of incorrect 2D tracklets tend to be mitigated, which means that, instead of rejecting an entire 2D tracklet, our method can locate and reject incorrect parts but keep the correct parts (\textit{sec.}~\ref{sec:link_sliding_windows})}. Moreover, when we aim to obtain 3D trajectories of body poses or 3D footprints on a stair, Type (1) methods may not work properly. Despite this defect, Type (1) methods provide an important insight: utilizing 2D tracklets other than single-frame 2D positions could form more robust features to verify cross-view consistency.
  
  Type (2) related works~\cite{chen2020multi,chen2020cross,ohashi20vmocap,dong2021fast} delve into the problem of tracking 3D pose trajectories more than 3D footprints, and most of them are online methods. \textit{To some extent, associating arbitrary positions (\eg, poses) is fundamentally different from associating footprints on the flat ground---it requires that multi-view stereo pairs are observable so that the 3D position can be calculated with triangulation}. To obtain multi-view stereo pairs, they take a single-frame multi-view association without temporal constraints. At every single frame, 3D positions are independently generated from single-frame 2D multi-view stereo pairs. Finally, the multi-frame association is employed to connect single-frame 3D positions to 3D trajectories. However, due to 2D-to-3D ambiguity and complicated human articulation, correctly associating multi-view 2D features in a single frame could be difficult.
  
  Taking the merits of Types (1) and (2), we introduce an online framework that combines the strengths and mitigates the limitations of the two approaches (see Table~\ref{tab:related_works}).  We inherit the idea of constructing 3D positions from paired cross-view positions using triangulation, but modifications are made by \textbf{collaboratively enforcing multi-frame multi-view association, 3D position calculation, and outlier rejection}. Our framework helps to enhance the robustness of 3D MM-Tracking with 2D bounding boxes and 2D poses as inputs.

 Notably, we skip the discussion of 3D MM-Tracking methods (\eg,~\cite{zhang2022voxeltrack}) that modulate the entire pipeline into neural networks. According to their settings, the weights of the neural network therefore encode the 3D structure of the reference views, and often cannot generalize to new scenes, which have different camera settings and 3D space. These methods may need to retrain their models to fit the new scene. Moreover, it is challenging to apply those methods to a large-scale space in the wild, since the entire observation space needs to be embedded to the neural network features. We aim to enable better generalization in our unified framework.

\begin{figure*}[!h]
  \centering
  \includegraphics[width=0.9\textwidth]{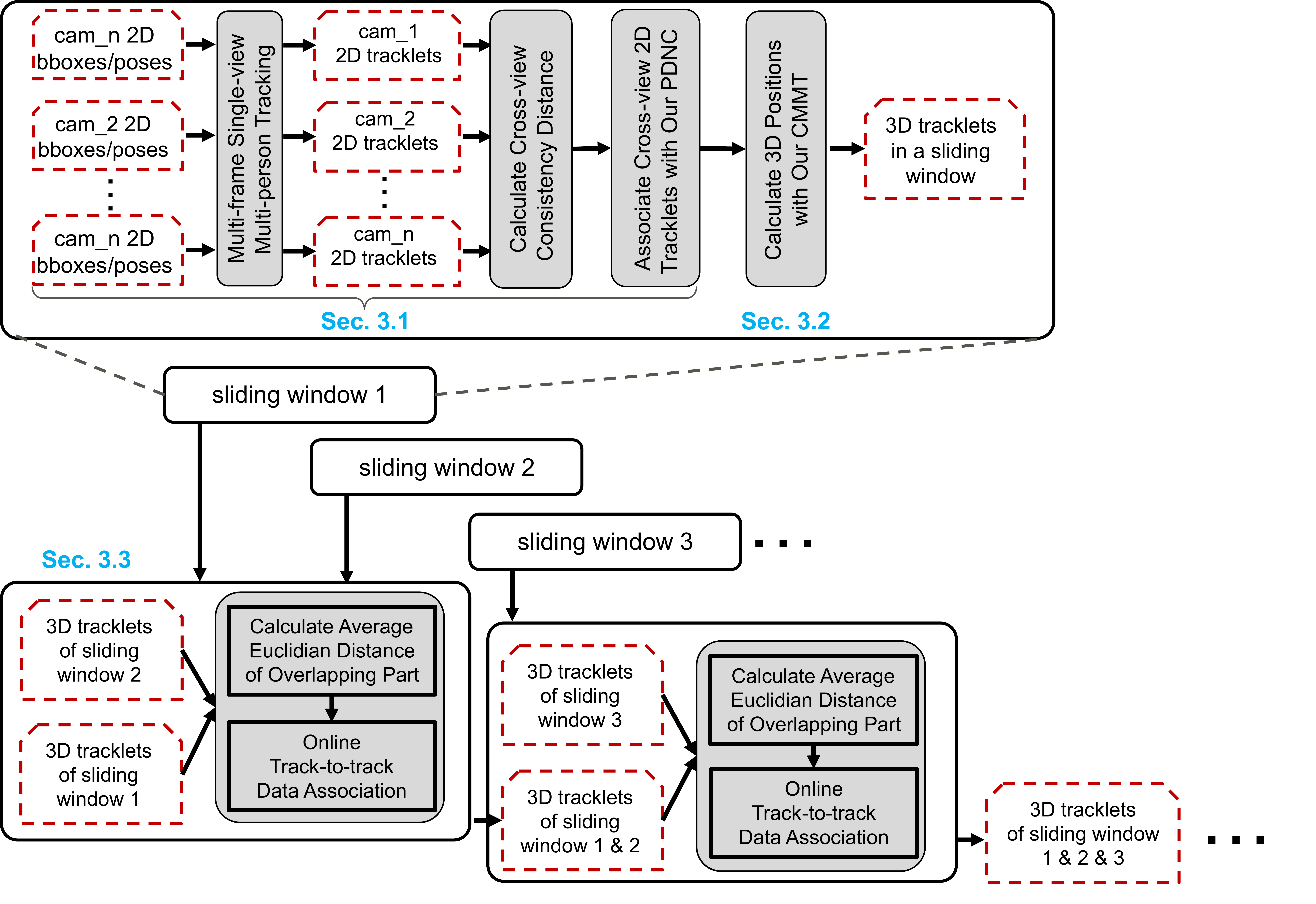}
  \caption{\textbf{Architecture of our proposed framework.} We utilize sliding windows to traverse the entire video. In each sliding window, Multiple-frame Single-view Multi-person Tracking is performed on each camera view to associate 2D positions into 2D tracklets. Then, multi-view 2D tracklets are associated to clusters using our \textbf{P}ropagable \textbf{D}istance-based \textbf{N}on-parametric \textbf{C}lustering (PDNC). Subsequently, we obtain the 3D positions by applying our \textbf{C}ollaborative \textbf{M}ulti-frame \textbf{M}ulti-view \textbf{T}riangulation (CMMT). Finally, we associate short-term 3D tracklets to long-term tracklets based on the Euclidian distance of their overlapping parts.
  }\label{fig:our_framework}
\end{figure*}

\section{Method}\label{sec:method}

Figure~\ref{fig:our_framework} illustrates the overall architecture of our framework. We further divided our framework into three phases: Collaborative Multi-frame Multi-View
Association (\textit{sec.}~\ref{sec:CMMA}), Collaborative Multi-frame Multi-view Triangulation (\textit{sec.}~\ref{sec:CMMT}), and Link Sliding Windows (\textit{sec.}~\ref{sec:link_sliding_windows}). The details of each phase are introduced in the following parts.

\subsection{Collaborative Multi-frame Multi-view Association}
\label{sec:CMMA}

  Presuming that multi-view videos are synchronized, the projective trajectories of a person in each view should be consistent in terms of positions and motions. Hence, multi-view 2D tracklets belonging to the same person can be matched by referring to the cross-view consistency. 
  
  We begin with a simple case to explain how to determine multi-view consistency. In a single frame, the distance of consistency between cross-view 2D positions is obtained by calculating the distance between one 2D position and the epipolar line constructed by the 2D position of another view. As introduced in previous works~\cite{canton2005towards, dong2019fast, chen2020cross}, such a calculation can be formulated as
\begin{equation}
  \begin{split} 
    d^{cross}(\boldsymbol{x}_{t,i}, \boldsymbol{x}_{t,j}) =& d\big(\boldsymbol{x}_{t,i}, \boldsymbol{l}_{c_{i}}(\boldsymbol{x}_{t,j})\big) + d\big(\boldsymbol{x}_{t,j}, \boldsymbol{l}_{c_{j}}(\boldsymbol{x}_{t,i})\big)\\
    \textrm{with}~ \boldsymbol{l}_{c_{i}} =& \bold{F}_{c_{j},c_{i}}\boldsymbol{x}_{t,j},\\
                  \boldsymbol{l}_{c_{j}} =& \bold{F}_{c_{i},c_{j}}\boldsymbol{x}_{t,i},\\
  \end{split} 
    \label{eq:D_epipolar_dist}
\end{equation}
 where $d^{cross}$ denotes the consistency distance between multi-view 2D positions $\boldsymbol{x}_{t,i}$ and $\boldsymbol{x}_{t,j}$ at frame $t$, and $d\big(\boldsymbol{x}_{t,i}, \boldsymbol{l}_{c_{i}}(\boldsymbol{x}_{t,j})\big)$ is the epipolar distance from the 2D position $\boldsymbol{x}_{t,i}$ to the epipolar line $\boldsymbol{l}_{c_{i}}(\boldsymbol{x}_{t,j})$. Furthermore, we denote $c_{i}$ as the camera ID of $\boldsymbol{x}_{t,i}$ and $\bold{F}_{c_{j},c_{i}}$ as the fundamental matrix maps a point from camera $c_{j}$ to $c_{i}$.

 \begin{figure}[!h]
  \centering
  \includegraphics[width=\linewidth]{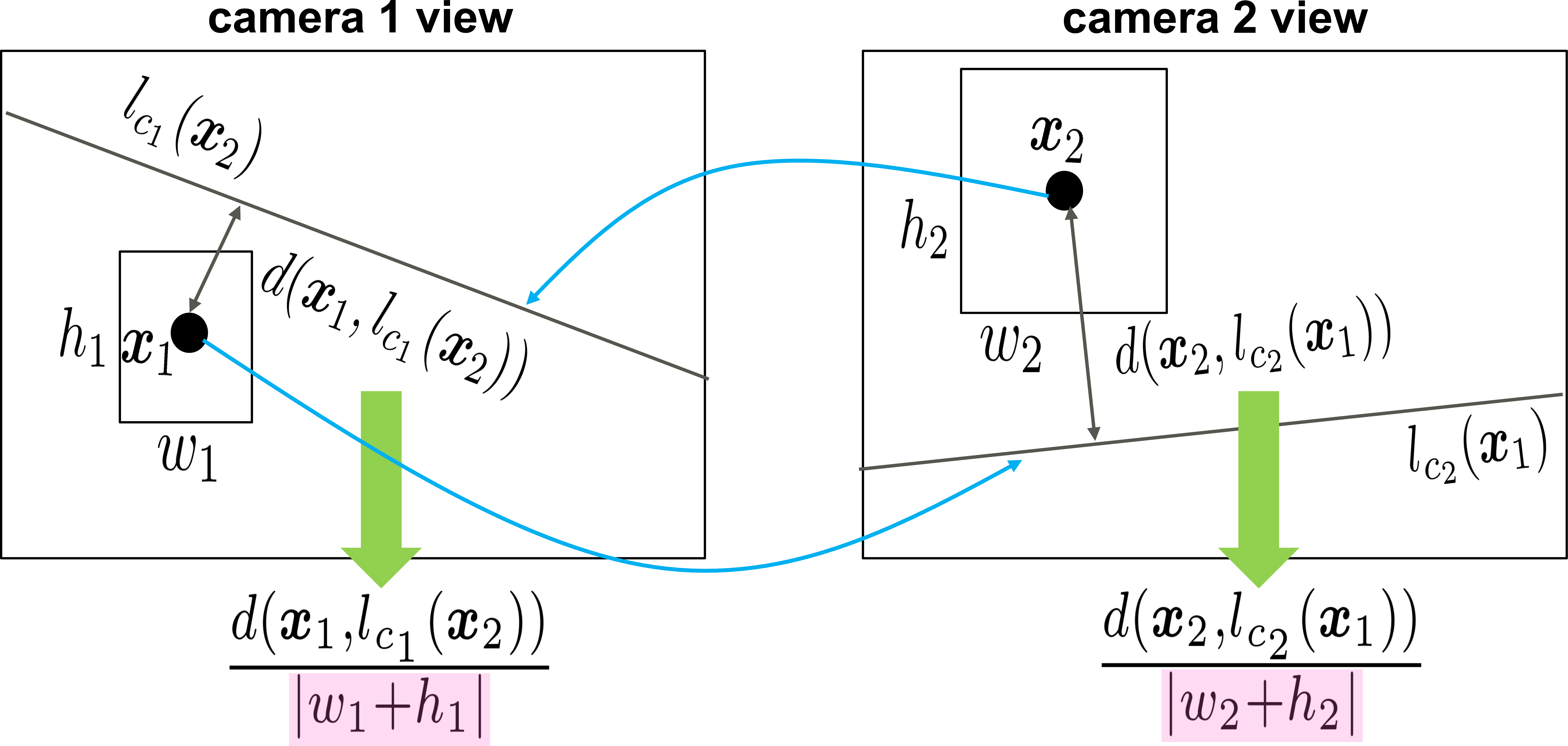}
  \caption{\textbf{Illustration of our normalized epipolar geometry distance.} We demonstrate an example by associating the bounding box center. Here, $w$ and $h$ denote the bounding box width and height, respectively, $p$ is the bounding box center coordinate; $l$ is the epipolar line and $d$ is the Euclidian distance from the target point to the given epipolar line. The epipolar distance has been normalized to be irrelevant to the 3D-to-2D projected variance caused by relative distance (see the highlighted parts). }\label{fig:norm_epipolar_distance}
  \end{figure}

  \begin{figure}[!h]
    \centering
    \includegraphics[width=\linewidth]{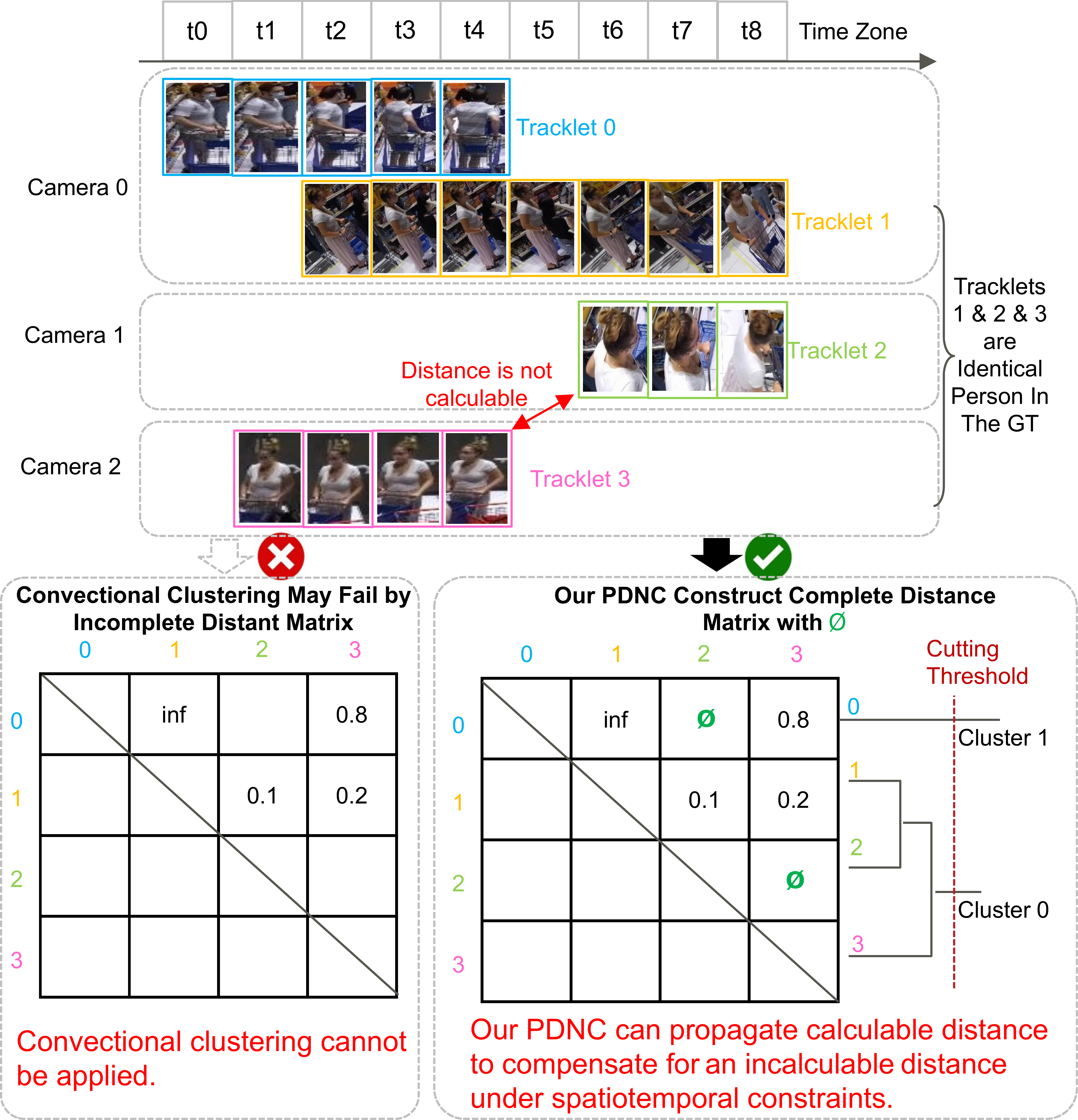}
    \caption{\textbf{Illustration of Distance Initialization in Propagable Distance-based Non-parametric Clustering (PDNC).} We streamline the cross-view association as a non-parametric clustering problem. From Eq.~\eqref{eq:multi-frame_2dx_distance_sets}, the cross-view consistency distance between tracklets $0$ and $1$ is set to be $\inf$ because they are in the same camera view and overlap in the temporal domain. The distance between tracklets $2$ and $3$ is incalculable since there is no temporal overlap between them. However, our PDNC can propagate a computable distance to compensate for the incalculable distance under spatiotemporal constraints.
    }\label{fig:PDNC}
  \end{figure}

 However, this formula does not account for the fact that the 2D projective scale of the same 3D distance possibly varies as a function of the object-to-camera distance. Because of the detection bias, the multi-view 2D positions may not match perfectly. Thus, their consistency distance is large for a small object-to-camera distance, and vice versa. Therefore, using Eq.~\eqref{eq:D_epipolar_dist} leads to two limitations. First, multi-view consistency can not be properly determined in a large viewing space, because the 2D projective scale divergence is notable. Second, the hyperparameter for multi-view association needs to be adjusted after the cameras have been relocated.  Although the previous method~\cite{dong2019fast} applied the mean and standard deviation of a batch of cross-view distances for normalization, their normalization seems to only reduce the inner batch variance other than the 3D-to-2D projected variance caused by the object-to-camera distance, which could increase the difficulty of selecting a constant threshold of epipolar distance for the cross-view association.

To solve this problem, it is prudent to normalize epipolar distance \textit{w.r.t.}~Eq.\ref{eq:D_epipolar_dist}, and we present the following modifications (\textit{c.f.} Figure~\ref{fig:norm_epipolar_distance}):
  \begin{equation} \label{eq:new_single_frame_epipolar_dist}
     \begin{split} 
    d^{cross}(\boldsymbol{x}_{t,i}, \boldsymbol{x}_{t,j}) =& \frac{d \big(\boldsymbol{x}_{t,i}, \boldsymbol{l}_{c_{i}}(\boldsymbol{x}_{t,j})\big)}{\textcolor{mypink1}{|w_{t,i}+h_{t,i}|}} + \frac{d\big(\boldsymbol{x}_{t,j}, \boldsymbol{l}_{c_{j}}(\boldsymbol{x}_{t,i})\big) }{\textcolor{mypink1}{|w_{t,j}+h_{t,j}|}},  \\
 \end{split}
  \end{equation}
where $w$ and $h$ are the horizontal and vertical scales of bounding boxes, respectively.

When 2D bounding boxes are utilized, we select their center point to calculate the person-to-person distance. When 2D poses are given, we generate 2D bounding boxes that encompass the 2D poses to obtain their scales. We also apply Eq.~\eqref{eq:new_single_frame_epipolar_dist} to each joint point and take their average value to represent the person-to-person distance. Unlike traditional approaches~\cite{he2020multi,chen2020multi,chen2020cross,ohashi20vmocap,dong2021fast}, in our approach, the epipolar distance has been normalized to be irrelevant to the 3D-to-2D projected variance caused by the object-to-camera distance.

\begin{algorithm*}[h!]
  \caption{\textbf{P}ropagable \textbf{D}istance-based \textbf{N}on-parametric \textbf{C}lustering (PDNC)}
  \label{alg:PDNC}
  \begin{algorithmic}[1]
  \begin{small}
    \Inputs{
      a) 2D tracklets cropped by the sliding window anchored at keyframe $k$. A 2D tracklet can be represented as $\mathcal{T}_{k,i}$, where $i$ is its index.\\
      b) Epipolar distance between multi-view 2D tracklets. Each of them can be represented as $d^{cross}_{k}(\mathcal{T}_{k,i}, \mathcal{T}_{k,j})$, where $i$ and $j$ are indices.
    }
    \Outputs{Final clustering set at key frame $k$, as $\Omega_{k}$.}

  \State \textbf{Initialize:} Assign each 2D tracklet to a cluster \strut$\omega_{k,i} \gets \mathcal{T}_{k,i}$, where $\omega_{k,i} \in \Omega_{k}$ and $i=1,\ldots,n$. And therefore $d^{cross}_{k}(\omega_{k,i}, \omega_{k,j})$ := $d^{cross}_{k}(\mathcal{T}_{k,i}, \mathcal{T}_{k,j})$.
    
  \While{A pair of clusters $\omega_{k,i}, ~\omega_{k,j} \in \Omega_{k}$, \textit{s.t.} $d^{cross}_{k} (\omega_{k,i}, ~\omega_{k,j}) < \lambda$} 
  
  \State a) Select two closest clusters $\omega_{k,i}$ and $\omega_{k,j}$.
  \State b) Merge two clusters to a new cluster $\omega_{k,p}  \gets \omega_{k,i} \cup \omega_{k,j}$. 
  \State c) Update distances from the new cluster $\omega_{k,p}$ to all other clusters in the current clustering set $\Omega_{k} $ by

  \For{$\omega_{k,q} \in \Omega_{k} $ and $p\neq q$}
   \State \textcolor{mygray}{\sout{$d^{cross}_{k}(\omega_{k,p}, ~\omega_{k,q}) = \text{max}\big(d^{cross}_{k} (\omega_{k,i}, ~\omega_{k,q}), d^{cross}_{k} (\omega_{k,j}, ~\omega_{k,q}) \big)$ } } \Comment{\textcolor{mygray}{Conventional Complete-linkage Clustering approach}} \\
  
   \[ ~~~
    \textcolor{mypink1}{
    d^{cross}_{k}(\omega_{k,p}, ~\omega_{k,q}) = 
    \begin{cases}
    \text{max}\big(d^{cross}_{k} (\omega_{k,i}, ~\omega_{k,q}), d^{cross}_{k} (\omega_{k,j}, ~\omega_{k,q}) \big),& ~\text{if}~ d^{cross}_{k} (\omega_{k,i}, ~\omega_{k,q}) ~\neq  \emptyset ~\text{and}~ d^{cross}_{k} (\omega_{k,j}, ~\omega_{k,q}) ~\neq  \emptyset\\
    d^{cross}_{k} (\omega_{k,j}, ~\omega_{k,q}),& ~\text{if}~ d^{cross}_{k} (\omega_{k,i}, ~\omega_{k,q}) = \emptyset ~\text{and}~ d^{cross}_{k} (\omega_{k,j}, ~\omega_{k,q}) ~\neq  \emptyset\\
    d^{cross}_{k} (\omega_{k,i}, ~\omega_{k,q}),& ~\text{if}~ d^{cross}_{k} (\omega_{k,j}, ~\omega_{k,q}) = \emptyset ~\text{and}~ d^{cross}_{k} (\omega_{k,i}, ~\omega_{k,q}) ~\neq  \emptyset\\
    \emptyset, & ~\text{if}~ d^{cross}_{k} (\omega_{k,i}, ~\omega_{k,q}) = \emptyset ~\text{and}~ d^{cross}_{k} (\omega_{k,j}, ~\omega_{k,q}) = \emptyset
    \end{cases} 
    }
    \] 
    \Comment{\textcolor{mypink1}{Our proposal}} 
 \EndFor
 \State d) Delete $\omega_{k,i}$ and $\omega_{k,j}$, as $\Omega_{k} = \Omega_{k} \setminus \left (\omega_{k,i}, \omega_{k,j}\right )$.

  \EndWhile
\end{small}
\end{algorithmic}
\end{algorithm*}

We have now introduced the consistency distance for multi-view 2D positions in a single frame. Most related works~\cite{chen2020multi,chen2020cross,dong2021fast,zhang2022voxeltrack} independently associate multi-view information in a single view, which could be suboptimal: different persons may have similar cross-view consistencies at a single frame, and the correct assignment may not always be established. We form a collaborative multi-frame multi-view association to improve the robustness of cross-view association. By jointly measuring the cross-view consistency in multiple frames, the cross-view inconsistency could be increased for different persons but decreased for identical persons. Consequently, a robust association can be achieved.

We are not the first to jointly perform multi-frame multi-view association. Before us, existing solutions~\cite{leal2012branch,zhang20204d} jointly optimized multi-frame multi-view association via 4D graphs formed with multi-frame multi-view information. However, their computational complexity increases dramatically when more frames are involved in 4D graphs. More specifically, considering that every two cross-frame cross-view detections can form an edge, the number of edges in their association graph could be calculated with the combination formula $\binom{N_{f}}{2}\binom{N_{c}}{2}\binom{N_{p}}{2}$, where $N_{f}$, $N_{c}$, and $N_{p}$ represents the number of frame, camera, and person, respectively. 

For videos with high recording rates, the inclusion of more frames enables the acquisition of more distinct motion information for a better multi-view association. Here, we consider an alternative formulation that ensures the efficient use of spatiotemporal information from arbitrary frames: through calculating the cross-view distance between 2D tracklets that involve multi-frame information, the computation required to jointly optimize multi-frame multi-view association is no longer tied to the complexity of frames, but it is only related to the number of multi-view 2D tracklets, thus improving its real-world applicability. Although we utilize information from multiple frames, the number of edges in our association graph is reduced to $\binom{N_{c}}{2}\binom{N_{p}}{2}$.

In our framework, we run multiple single-camera MOT trackers simultaneously (\eg, SORT~\cite{Bewley2016_sort}) to obtain 2D tracklets for each camera view.
To achieve parallel single-camera tracking, we utilize sliding windows to pass through the obtained 2D tracklets. Within a sliding window whose center is anchored at keyframe $k$ (\textit{c.f.} Figure~\ref{fig:our_framework}), the distance between cross-view 2D tracklets becomes a set of normalized epipolar distances. Supposing we have obtained 2D tracklets $\mathcal{T}_{k,i}$ and $\mathcal{T}_{k,j}$, which are from cameras $c_i$ and $c_j$, respectively, we represent such a distance set with $S$ and formulate it as follows:
\begin{equation}\label{eq:multi-frame_2dx_distance_sets}
\begin{split} 
  \mathcal{S}(\mathcal{T}_{k,i}, \mathcal{T}_{k,j}) :=& 
    \big\{d^{cross}(\boldsymbol{x}_{t,i}, \boldsymbol{x}_{t,j}) \mid  t \in \Psi_{i} \cap \Psi_{j} \big\}, \\
\end{split}
\end{equation}
where $\Psi_{i}$ and $\Psi_{j}$ denote the frames covered by  $\mathcal{T}_{k,i}$ and $\mathcal{T}_{k,j}$, respectively.

Then, we specifically design the distance between $\mathcal{T}_{k,i}$ and $\mathcal{T}_{k,i}$ based on three conditions. When $\mathcal{T}_{k,i}$ and $\mathcal{T}_{k,i}$ have temporal overlap and their camera IDs are different, we take the mean of $S(\mathcal{T}_{k,i}, \mathcal{T}_{k,j})$. When there is no temporal overlap, regardless of whether their camera IDs are different or identical, we assign  $\emptyset$ to the distance. Here, $\emptyset$ serves as a special symbol that will trigger the merge conditions in our PDNC (see Figure~\ref{fig:PDNC} and Algorithm~\ref{alg:PDNC}). When temporal overlap exists and the camera IDs are the same, we set an infinitely large value and conjecture that these 2D tracklets should be assigned to different persons. The corresponding formula is expressed as follows:
\begin{small}
\begin{equation}\label{eq:tracklets_dist}
  \begin{split}  
    d^{cross}_{k}(\mathcal{T}_{k,i}, \mathcal{T}_{k,j}) =~~~~~~~~~~~~~~~~~~~~~~~~~~~~~~~~~~~~~~~~~~~~~~~~~~~~~~~~~~~~~~  \\
  \left\{\begin{matrix}
    & \textrm{mean} \big( S \left(\mathcal{T}_{k,i}, \mathcal{T}_{k,j}\right) \big), & \text{if} & \Psi_{i} \cap  \Psi_{j} \neq \emptyset ~\text{and}~ c_{i} \neq c_{j},\\ 
    & \textcolor{mypink1}{\emptyset,} & \textcolor{mypink1}{\text{if}} & \textcolor{mypink1}{\Psi_{i} \cap  \Psi_{j} =  \emptyset ,}\\
    & \textcolor{mypink1}{\inf,} & \textcolor{mypink1}{\text{if}} & \textcolor{mypink1}{ \Psi_{i} \cap  \Psi_{j} \neq \emptyset~\text{and}~c_{i} = c_{j} .}
  \end{matrix}\right. 
\end{split}
\end{equation}
\end{small}

With the aforementioned cross-view consistency distance, associating cross-view 2D tracklets can be formulated as a clustering problem and aim to optimize global criteria. Although clustering methods have been widely applied in tracking tasks, we should be aware of three issues in associating cross-view 2D tracklets. First, predefining the number of clusters or even specifying the maximum size of clusters (\eg,~\cite{he2020multi}) should be avoided; otherwise, the framework may degenerate if more observed persons are included. We tend to apply non-parametric clustering methods~\cite{roberts1997parametric} to automatically learn the number of clusters from data. Second, most non-parametric clustering methods contain abstract hyperparameters that need to be interpreted; thus, it is difficult to use them in real applications. The hyperparameter of Complete-linkage Clustering is a distance threshold that has an intuitive physical definition. We take Complete-linkage Clustering as a candidate. Nonetheless, in contrast to single-frame multi-view association~\cite{dong2019fast,chen2020cross}, we face the challenges posed by multi-frame multi-view association: \textbf{when multi-view 2D tracklets have no overlapping frames, their distance cannot be directly calculated, therefore, conventional Complete-linkage Clustering, which requires distance of all elements in clustering, cannot be applied in this scenario}. To tackle this issue, we proposed a novel \textbf{P}ropagable \textbf{D}istance-based \textbf{N}on-parametric \textbf{C}lustering (PDNC) and describe its details in Algorithm~\ref{alg:PDNC}. In PDNC, we substitute distance updating in conventional Complete-linkage Clustering with a
dynamic updating mechanism. Utilizing our distance definition in Eq.~\eqref{eq:tracklets_dist}, PDNC propagates the calculable distance to compensate for the incalculable distance under spatiotemporal constraints.

\subsection{Collaborative Multi-frame Multi-view Triangulation}
\label{sec:CMMT}

\begin{algorithm*}[h!]
  \caption{Collaborative Multi-frame Multi-view Triangulation}
  \label{alg:CMMT}
  \begin{algorithmic}[1]
    \begin{small}
    \Inputs{
      2D tracklets cropped by the sliding window anchored at keyframe $k$ and assigned to cluster $p$ by using Algorithm~\ref{alg:PDNC}. A 2D tracklet is represented as $\mathcal{T}_{k,p,i}$, where $i$ is its index. We have $\mathcal{T}_{k,p,i} \in \omega_{k,p}$ and $\omega_{k,p} \in \Omega_{k}$.}
    \Outputs{3D tracklets anchored at keyframe $k$ and assigned to cluster $p$. A 3D tracklet is denoted as $\mathcal{U}_{k,p}$.}
  \State \textbf{Initialize:} Copy each 2D tracklets to observed 2D tracklets, as $\mathcal{T}^{obs}_{k,p,i} \gets \mathcal{T}_{k,p,i}$, $\forall$  $i=1,\cdots,n$. 
  \State Apply interpolation function $f(\cdot)$ (\eg, linear interpolation) to observed 2D tracklets to get interpolated 2D tracklets, as $\mathcal{T}^{infill}_{k,p,i} = f \big(\mathcal{T}^{obs}_{k,p,i} \big)$, $\forall$  $i=1,\cdots,n$. \Comment{\textcolor{mygray}{Multi-frame information is implicitly involved through interpolation}} 
  \State Merge observed and interpolated 2D tracklets with identical indices, as $\mathcal{T}^{merge}_{k,p,i} \gets \mathcal{T}^{obs}_{k,p,i} \cup \mathcal{T}^{infill}_{k,p,i}$, $\forall$ $i=1,\cdots,n$. The multi-view relationship remains the same: $\mathcal{T}^{merge}_{k,p,i} \in \omega_{k,p}$ and $\omega_{k,p} \in \Omega_{k}$.  

  \For{ $\omega_{k,p}$ in $\Omega_{k}$ }
  \For{multi-view stereo pairs formed in each frame of $\mathcal{T}^{merge}_{k,p,i}$, where $\mathcal{T}^{merge}_{k,p,i} \in \omega_{k,p}$ }
    \State a) Apply Eq.~\eqref{eq:3d_triangulation} to multi-view stereo pairs to yield a set of 3D candidate positions at one frame, as $\Pi$.
    \State b) Apply Complete-linkage Clustering on $\Pi$ with cutting threshold $\kappa$.
    \State c) Select the largest cluster at each frame and calculate their mean value for 3D position $\boldsymbol{X}_{k,p}$ with Eq.~\eqref{eq:fuse_largest}.
    \State d) Add $\boldsymbol{X}_{k,p}$ to 3D tracklet $\mathcal{U}_{k,p}$.
  \EndFor
  \EndFor
\end{small}
\end{algorithmic}
\end{algorithm*}

After we obtain the multi-view relationship, 3D positions are inferred from multi-view 2D positions by applying linear algebraic triangulation~\cite{andrew2001multiple}. Suppose that we have a pair of cross-view 2D positions that can be represented in homogeneous coordinates as $\boldsymbol{x}_{i} = (x_{i},y_{i},1)$ and $\boldsymbol{x}_{j} = (x_{j},y_{j},1)$, where $x$ and $y$ are the position values on an image coordinate system. Their corresponding perspective projection matrices are $\bold{P}_{c_i}$ and $\bold{P}_{c_j}$, respectively. In general, the 3D position $\boldsymbol{\tilde{X}}$ is estimated by solving the following equations:
%\begin{small}
  \begin{equation}\label{eq:3d_triangulation}
  \begin{split} 
    \mathbf{A} \boldsymbol{\tilde{X}}=& ~ 0, \\
    \textrm{with}~\mathbf{A} = &
\begin{bmatrix}
  x_{i}\bold{P}_{c_i, 3} - \bold{P}_{c_i,1}\\
  y_{i}\bold{P}_{c_i,3} - \bold{P}_{c_i,2}\\
  x_{j}\bold{P}_{c_j, 3} - \bold{P}_{c_j,1}\\
  y_{j}\bold{P}_{c_j,3} - \bold{P}_{c_j,2}\\
 \end{bmatrix},
  \end{split}
  \end{equation}
%\end{small}
where $\bold{P}_{c_i, n}$ is the $n$-th row of $\bold{P}_{c_i}$~\cite{andrew2001multiple}.

If we equally treat 2D positions in the triangulation approach, then the
estimation errors of 2D positions and incorrect cross-view association could impair the accuracy of 3D positions. Toward the goal of reducing estimation errors, Iskakov \etal~ first proposed to apply the Random Sample Consensus (RANSAC)~\cite{fischler1981random} 
%together with the Huber loss~\cite{huber1992robust} 
to remove outliers, and further recommended a volumetric triangulation approach to jointly employ inlier and outlier information to generate improved 3D positions~\cite{iskakov2019learnable}. Due to its simplicity, similar solutions have been applied in other works. Nonetheless, such a solution only considers the information in a single frame, while spatiotemporal information of nearby frames is ignored. Moreover, RANSAC can only handle a moderate percentage of outliers without the cost blowing up~\cite{lowe2004distinctive}, but we may have a high rate of outliers caused by association failures and estimation errors.

\begin{figure}[!h]
  \centering
  \includegraphics[width=\linewidth]{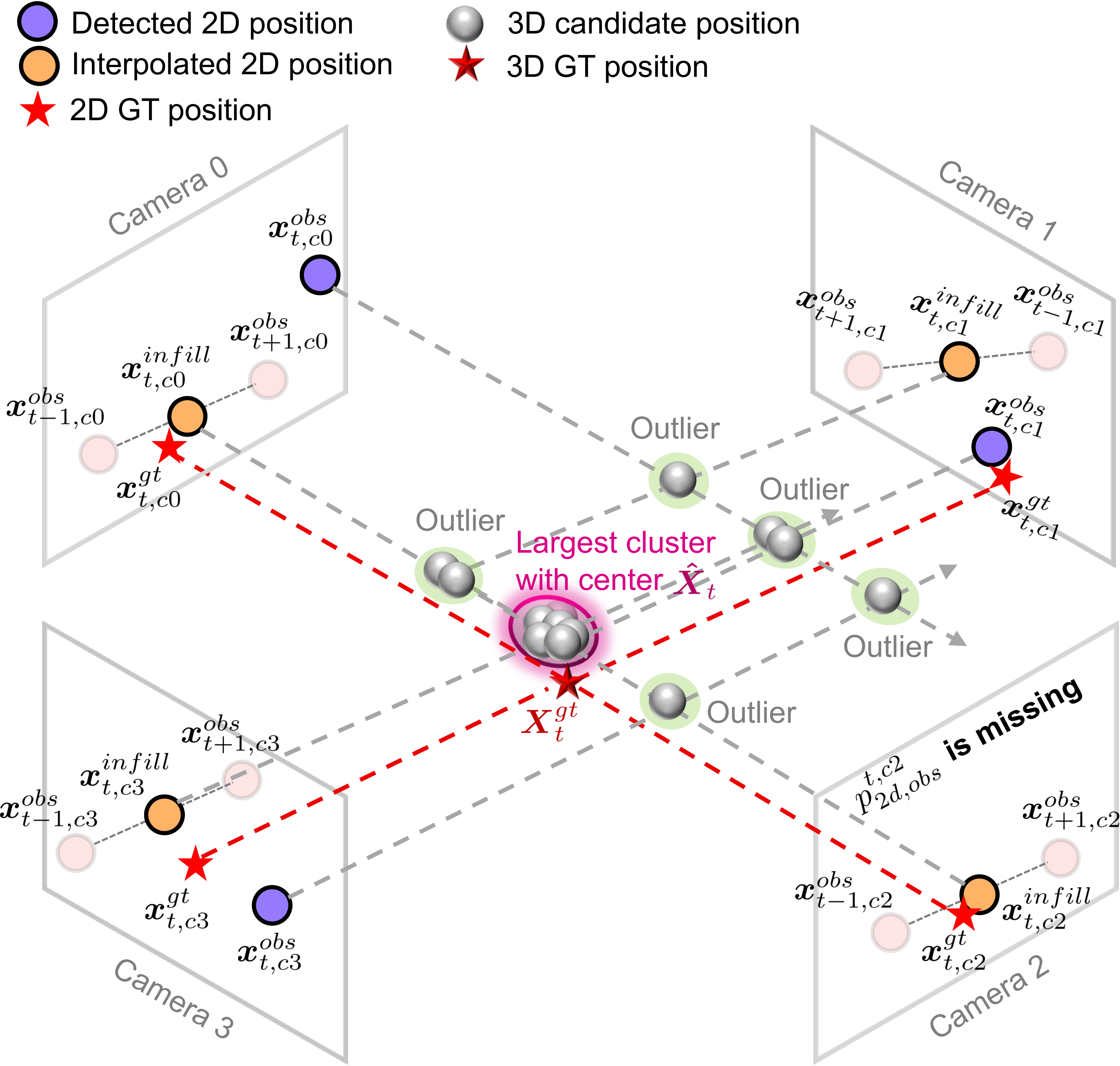}
  \caption{\textbf{Illustration of \textbf{C}ollaborative \textbf{M}ulti-frame \textbf{M}ulti-view \textbf{T}riangulation (CMMT).} Suppose both observed 2D position $\boldsymbol{x}^{obs}_{t}$ and interpolated 2D position $\boldsymbol{x}^{infill}_{t}$ could be incorrect in a single monocular view (\ie, different from ground truth $\boldsymbol{x}^{gt}_{t}$). After clustering their constructed 3D candidate positions (gray circles), we use the centroid of the largest cluster to approximate the 3D ground truth $\boldsymbol{X}^{gt}_{t}$ based on the majority vote. 
  }\label{fig:CMMT}
\end{figure}

We assume that the spatiotemporal information of nearby frames can be used to obtain the correct 3D positions. Consequently, a novel \textbf{C}ollaborative \textbf{M}ulti-frame \textbf{M}ulti-view \textbf{T}riangulation (CMMT) is proposed to integrate the spatiotemporal information to 3D position calculation.  The details of CMMT are shown in Figure~\ref{fig:CMMT} and explained in Algorithm~\ref{alg:CMMT}. Algorithm~\ref{alg:PDNC} is applied for the association, whereas, Algorithm~\ref{alg:CMMT} is designed for outlier rejection and 3D position calculation. Although both of them have jointly utilized multi-frame multi-view information and applied non-parametric clustering, their functions are fundamentally different.

In CMMT, we adopt a simple approach---interpolation---to encode multi-frame information to interpolated 2D tracklets. Supposing that the interpolation window size is $\varphi$ and the 1D interpolation function is $f(\cdot)$ (\eg, linear interpolation), we can generate the interpolated 2D position $\boldsymbol{x}^{infill}_{i}$ using the observed 2D positions $\boldsymbol{x}^{obs}_{i}$, which is formulated as follows
\begin{equation}\label{eq:2d_interpolation}
\boldsymbol{x}^{infill}_{i} = f(\boldsymbol{x}^{obs}_{i}), ~~\forall~ i = i-\varphi ,\cdots, i+\varphi 
\end{equation}

Applying Eq.~\eqref{eq:2d_interpolation} to each element of the 2D tracklet $\mathcal{T}^{obs}_{i}, ~\forall~ i = 1,\cdots, n$, we generate interpolated 2D tracklets $\mathcal{T}^{infill}_{i},~\forall~ i = 1,\cdots, n$. Then, we merge $\mathcal{T}^{obs}_{i}$ and $\mathcal{T}^{infill}_{i}$ into $\mathcal{T}^{merge}_{i},~\forall~ i = 1,\cdots, n$. A rich amount of 3D candidate positions, as a set  $ \Pi = \{\boldsymbol{X}^{cand.}_{i},\cdots,\boldsymbol{X}^{cand.}_{m}\}$, could be generated by applying Eq.~\eqref{eq:3d_triangulation} to $\mathcal{T}^{merge}_{i},~\forall~ i = 1,\cdots, n$. 

Nonetheless, the 3D positions generated from interpolations could either compensate for the missing observations as inliers or delude existing observations as outliers. Thus, we apply the conventional Complete-linkage Clustering~\cite{murtagh2012algorithms} to identify the pattern of 3D candidate positions, as a set of clusters $\{{\omega_{1}, \cdots, \omega_{p}}\} \in \Omega$. Then, we regard the largest cluster $\omega^{largest}$ as the inlier; meanwhile, the others are treated as outliers. The final 3D position is obtained from the centroid of the inlier. 

\begin{equation}\label{eq:fuse_largest}
  \begin{split} 
  \boldsymbol{X} =& \frac{1}{\left | \omega^{largest} \right |} \sum _{i \in \omega^{largest}} \boldsymbol{X}^{cand.}_{i},\\
  \mathrm{s.t.}~  & \lVert \boldsymbol{X} - \boldsymbol{X}^{cand.}_{i} \rVert \leqslant \kappa,
  \end{split} 
  \end{equation}
where  $\kappa$ denotes the cutting threshold for the clustering.

Our CMMT still works when the percentage of inliers is less than $50\%$, which may be difficult for other outlier rejection methods. In terms of computational complexity, our PDNC and CMMT are modified from the complete-linkage clustering, which has a time complexity of $\mathcal{O} \left ( n^{2} \right )$, where $n$ is the number of elements to be clustered.  In the MM-Tracking, if each camera can capture all persons, the number of cameras and persons are $N_{c}$ and $N_{p}$, respectively, and we have $n= N_{p} N_{c} $, so the time complexity of our PDNC and CMMT can be approximated by $\mathcal{O} \left ( N_{p}^{2} N_{c}^{2} \right )$. However, in our test datasets, each camera only captures a few persons; moreover, since we assume that observations in the same camera view should not be clustered, we skip the distance calculation and clustering process for observations in the same camera view. Therefore, the actual complexity can generally be less than $\mathcal{O} \left ( N_{p}^{2} N_{c}^{2} \right )$. In practice, the number of cameras and people observed may not be too large, so the overall computational cost could be acceptable.

\subsection{Link Sliding Windows}
\label{sec:link_sliding_windows}
As illustrated in Figure~\ref{fig:our_framework}, our previous processes have generated short-term 3D tracklets within a sliding window. In this subsection, we will learn how to link short-term 3D tracklets to the final result with our online track-to-track data association.

In conventional track-to-track data association works~\cite{chen2020multi,chen2020cross,ohashi20vmocap,dong2021fast}, the cross-camera relationship is unknown and there exist multiple redundant 3D tracklets for identical persons. Thus, those methods are mainly used to obtain the cross-camera relationship with a clustering approach. In our previous processes, however, the cross-camera relationship has been identified in each sliding window. Applying the conventional track-to-track methods discards our obtained cross-camera relationship and makes our previous effects useless.

To fully leverage the cross-camera relationship obtained in each sliding window, we propose an online track-to-track data association. The short-term 3D tracklets in each sliding window are treated as nodes in our online track-to-track data association. We apply linear assignment to associate these nodes across sliding windows in 3D space. Tracklets between two adjacent sliding windows should be connected if they are compatible. Based on the track management of SORT~\cite{Bewley2016_sort}, we form a track-to-track management to decide how and when to initialize, update and terminate a long-term tracklet that is connected from short-term tracklets.

%As illustrated in Figure~\ref{fig:our_framework}, our previous processes have generated short-term 3D tracklets within a sliding window. In this subsection, we will learn how to link short-term 3D tracklets to the final result.

%We perform a track-to-track data association to link short-term to long-term 3D tracklets. Note that, \textit{our track-to-track data association is different from those used in previous works}~\cite{chen2020multi,chen2020cross,ohashi20vmocap,dong2021fast}. In previous works~\cite{chen2020multi,chen2020cross,ohashi20vmocap,dong2021fast}, the cross-camera relationship is unknown before their track-to-track data association and has multiple redundant 3D tracklets for identical persons. Thus, they generally utilized clustering methods to fuse 3D tracklets in their track-to-track data association. In our approach, however, the cross-camera relationship is identified in each sliding window. Thus, using the cross-camera relationship, a unique short-term 3D tracklet is generated for each person. As a novel approach, we treat each short-term 3D tracklet as a unit and perform the conventional single-view MOT to associate those units in the 3D space. 

More specifically, for two 3D tracklets $\mathcal{U}_{k,i}$ and $\mathcal{U}_{k+1,j}$, with centers anchored at keyframes $k$ and $k+1$, respectively, we set their distance as the average Euclidian distance between their overlapping parts. The distance matrix of 3D tracklets between two adjacent sliding windows is formulated as
  \begin{equation}\label{eq:3d_euclidian_dist}
  \begin{split} 
    \mathbf{D}=&\{D_{i,j}\},\\
    D_{i,j} =& \frac{1}{\left |\Psi_{i}\cap \Psi_{j}\right |}\sum
    \lVert \boldsymbol{X}_{t,m} - \boldsymbol{X}_{t,n} \rVert,\\
    \mathrm{s.t.}~ &~t\in \Psi_{i}\cap \Psi_{j},~ \boldsymbol{X}_{t,m} \in \mathcal{U}_{k,i},~ \boldsymbol{X}_{t,n} \in \mathcal{U}_{k+1,j}
  \end{split}
  \end{equation}
  where $\left |\Psi_{i}\cap \Psi_{j}\right |$ denotes the number of overlapping frames between $\mathcal{U}_{k,i}$ and $\mathcal{U}_{k+1,j}$, and $\boldsymbol{X}_{t,m}$ and $\boldsymbol{X}_{t,n}$ are the 3D points in $\mathcal{U}_{k,i}$ and $\mathcal{U}_{k+1,j}$, respectively.

  \begin{figure}[!h]
    \centering
    \includegraphics[width=\linewidth]{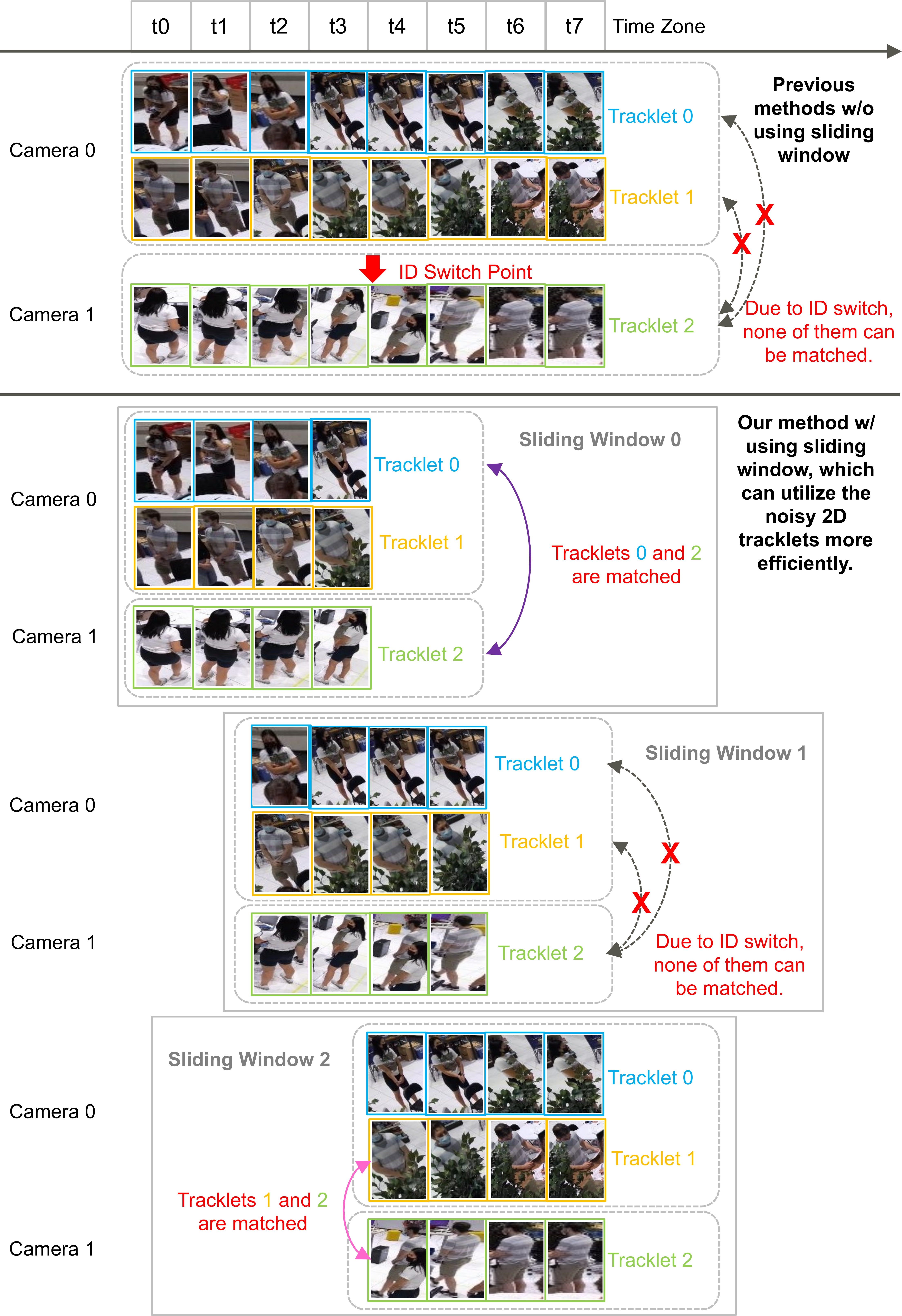}
    \caption{\textbf{Our sliding-window approach can utilize the noisy 2D tracklets more efficiently.} Previous methods~\cite{sternig2011multi,wen2017multi,kohl2020mta, he2020multi}, that take the entire 2D tracklets for the cross-view association, may not be able to utilize 2D tracklets if ID Switches are encountered. Our method uses the sliding windows to alleviate this problem.
    }\label{fig:sw_metrit}
  \end{figure}

  Following the most fundamental strategy of single-view MOT~\cite{Bewley2016_sort, Wojke2017simple}, we formulate the 3D tracklets connection problem as a linear assignment problem. The relationship between the 3D tracklets of two adjacent sliding windows is modeled as a bipartite graph, which can be further represented as a Boolean matrix $\mathbf{M}$. When row $i$ is assigned to column $j$, we have $\mathbf{M}_{i,j}=1$. Each row is assigned to at most one column, and each column is assigned to at most one row. Thus, the optimal assignment $\mathbf{M}^*$ is obtained by minimizing the total cost, as follows
  \begin{equation}\label{eq:hungarian}
    \begin{split} 
  \mathbf{M}^* =& \argmin_{\mathbf{M}} \sum_{i,j} \mathbf{D}_{i,j}  \mathbf{M}_{i,j}\\
  \textrm{subject to} &\sum_{i} \mathbf{M}_{i} = 1  \\
                      &\sum_{j} \mathbf{M}_{j} = 1 . 
    \end{split}
  \end{equation}
  
  Referring to $\mathbf{M}^*$, the matched 3D tracklets of the sliding window $k+1$ are linked to the corresponding 3D tracklets of the sliding window $k$, whereas the unmatched 3D tracklets of the sliding window $k+1$ are regarded as new tracklets. After passing through all sliding windows, short-term 3D tracklets are linked to our final results.

  Except for approaching online processing, another merit of using the sliding window for 3D MM-Tracking is illustrated in Figure~\ref{fig:sw_metrit}. Previous methods~\cite{sternig2011multi,wen2017multi,kohl2020mta, he2020multi}, that take the entire 2D tracklets for the cross-view association, may discard a 2D tracklet if ID Switches are encountered. Nonetheless, rejecting 2D tracklets generally have poor tracking capability in areas captured by a limited number of cameras. To alleviate this problem, our method uses the sliding window. 2D tracking errors could be limited to one sliding window and do not affect the others.
  The demand for online processing and the effective use of tracking features have been successfully realized in our framework by adapting sliding windows in 3D-MM tracking.

\section{Experiments}

\subsection{Evaluation Datasets}
We performed experiments on four public datasets. Among them, the \textit{Campus} and \textit{Shelf}~\cite{fleuret2007multicamera,BelagiannisAASN16} datasets focus on 3D pose tracking, whereas the \textit{WILDTRACK}~\cite{chavdarova2018wildtrack} and \textit{MMPTRACK}~\cite{han2021mmptrack} datasets focus on 3D footprint tracking.

The \textit{Campus} and \textit{Shelf}~\cite{fleuret2007multicamera, BelagiannisAASN16} datasets provide calibrated camera parameters and videos for multi-camera multi-person 3D pose tracking.
While the \textit{Campus} dataset consists of three cameras and three persons, the \textit{Shelf} dataset includes five cameras and four persons. To infer the 3D pose trajectories above the ground plane, the methods used in previous works~\cite{sternig2011multi,wen2017multi,kohl2020mta,he2020multi} cannot be applied, since their 3D positions are computed using the homography projection matrix from the camera view to the ground plane.

The \textit{WILDTRACK}~\cite{chavdarova2018wildtrack} dataset captures the 3D footprint of $313$ individuals on a planar ground, by using seven calibrated cameras. Due to the large number of individuals, their projections in each camera view are heavily occluded, and therefore it is challenging to associate their cross-view observations in a single frame. Our framework jointly leverages multi-frame multi-view information to make robust tracking on this dataset. In our experiments, we followed the setup of LMGP~\cite{nguyen2022lmgp} to apply the first $360$ frames for training and the rest for testing. We also utilize the same detections as LMGP~\cite{nguyen2022lmgp} did.

\begin{table*}[!hbt]
  \centering
  \setlength{\tabcolsep}{2pt}
  \caption{\textbf{Comparison with the state-of-the-art methods on the \textit{WILDTRACK} dataset~\cite{chavdarova2018wildtrack}}. MOTA is used as the dominant evaluation metric. The definitions of MOTA and IDF1 are provided in \textit{refs.}~\cite{MOTChallenge2015,milan2016mot16}. The data rendered in \textbf{Bold} and \underline{Underlined} indicate the best and second-best results respectively. }
  \label{tab:multi-wildtrack}
  \begin{tabular}{cccccccccc}
  \toprule
  Method  & IDF1 $\uparrow$ & MOTA  $\uparrow$ & MT  $\uparrow$   & ML  $\downarrow$  & FP $\downarrow$ & FN $\downarrow$ & IDs $\downarrow$ &\makecell[c]{Using appearance \\feature  w/ \\Re-ID model} & Approach\\ 
  \midrule
  Chavdarova \etal~\cite{chavdarova2018wildtrack} (CVPR 18)   & 78.4\%  & 72.2\%   & 42.1\% & 14.6\% & 2007& 5830& 103 &\cmark &Offline\\ 
   BaTuong \etal~\cite{OngVVKN20DBLP} (TPAMI 20)   & 72.5\%  & 70.1\%   & \underline{93.6\%} & 22.8\% & 960 & 990 & 107 &\xmark & Online \\ 
  You \etal~\cite{you2020real} (ArXiv 20)  & 81.9\%  & 74.6\%   & 65.9\%  & 4.9\%  & 114 & 107 & 21  &\xmark &Online \\ 
  Nguyen \etal~\cite{nguyen2022lmgp} (CVPR 22) & \textbf{98.2\%} & \textbf{97.1\%}  & \textbf{97.6\%} & \textbf{1.3\%} & 71 & \textbf{7}  & 12 & \cmark &Offline  \\ 
  \textbf{Ours (w/o appearance)}  & 94.7\%  & 94.0\%  & 87.8\%  & 7.3\%  & \textbf{22} & 24& \underline{11}  & \xmark &Online \\ 
  \textbf{Ours (w/ appearance)}  & \underline{97.5\%}  & \underline{95.2\%}   & 92.7\%  & \underline{2.4\%}  & \underline{33} & \underline{8} & \textbf{5}  & \cmark &Online \\ 
  \bottomrule
  \end{tabular}
\end{table*}

The \textit{MMPTRACK}~\cite{han2021mmptrack} dataset is the \textbf{largest} publicly available multi-view multi-person tracking dataset, which was introduced in the ICCV 2021 Multi-camera Multiple People Tracking Challenge. In total, more than $9.6$ hours of videos were collected, covering $28$ subjects in five representative environments: Retail, Lobby, Industry, Cafe, and Office. Depending on the environment, the number of cameras varies from four to six. The dataset is divided into training, validation, and test sets. While the annotations of the training and validation sets are given, the ground truth of the test set is hidden from the public for a fair comparison. In our evaluation, we submitted our tracking results to the official evaluation server
%~\footnote{\url{https://competitions.codalab.org/competitions/33729}} 
to obtain the performance.

\subsection{Evaluation Metrics}
In our experiments, we jointly evaluate the tracking performance and 3D pose estimation performance. We follow the conventional setting to employ CLEAR metrics (\eg, MOTA)~\cite{bernardin2008evaluating} and Identity metrics (\eg, IDF1)~\cite{ristani2016performance} to evaluate the tracking results from various perspectives. In detail, IDs (\ie, Number of ID switches) indicate the times of identity jumps, IDF1 (\ie, ID F1 scores) accounts for identity match performance, and MOTA (\ie, Multi-Object Tracking Accuracy) is a combination of false positives, missed targets and IDs. Besides, MT (\ie, number of mostly tracked trajectories) measures the number of trajectories whose target was tracked more than $80\%$, whereas ML (\ie, number of mostly lost trajectories) measures the number of trajectories that have less than $20\%$ target tracked. In addition, FP and FN represent the ratio of False Positives and False Negatives, respectively. Among them, the MOTA score is the dominant metric used to measure the overall tracking performance. To evaluate the 3D pose estimation performance, we follow the trends to apply the PCP (\ie, Percentage of Correctly estimated Parts), which is described in the \textit{Campus} and \textit{Shelf}~\cite{fleuret2007multicamera, BelagiannisAASN16} datasets.

\subsection{Experiment Setup}
Since detection and tracking are separated in our framework, we decouple their settings as follows.
For detection, we choose the off-the-shelf object detector---YOLOX~\cite{yolox2021}---to train and inference bounding boxes for our experimental datasets. We follow the default settings of YOLOX and run it on a single GPU machine (NVIDIA A100). For the tracking, we set up hyperparameters of Table~\ref{tab:notations} as $\nu=30,\delta=20,\lambda=0.3, \varphi=7, \kappa=0.2$.
In addition, although our suggested framework is appearance-free, we also perform the ablation study to investigate the effect of using the appearance feature. To obtain the appearance feature, we utilize the model of Strong-ReID~\cite{Luo_2019_Strong_TMM} with one GPU machine (NVIDIA A100). Then we follow the strategy of DeepSORT~\cite{Wojke2017simple} to fuse the geometry and appearance features in the single-view and multi-view data association.

\begin{table*}[t]
  \begin{center}
  \setlength{\tabcolsep}{0.2pt}
  %\footnotesize
  \caption{\textbf{Evaluation result on the \textit{MMPTRACK} Dataset~\cite{han2021mmptrack}.}  MOTA is used as the dominant evaluation metric. The definitions of MOTA and IDF1 are provided in \textit{refs.}~\cite{MOTChallenge2015,milan2016mot16}. The data rendered in \textbf{Bold}  indicates the best results. } \label{tab:MMPTRACK}
  \begin{tabular}{lccccc}
    \toprule
    Rank & MOTA$\uparrow$ & IDF1$\uparrow$  & \makecell[c]{Using appearance \\feature  w/ \\Re-ID model}& \makecell[c]{\textbf{Access test set} \\w/ Semi-supervised\\ Learning } &\makecell[c]{Average \\Speed $\uparrow$}\\
    \midrule
    \multicolumn{6}{l}{Before challenge deadline submissions} \\
    
   $1^{st}$ (Alibaba DAMO Academy) & \textbf{96.0\%} & \textbf{97.6\%}  & \cmark & \cmark& $<$  1 FPS\\
   $2^{nd}$ (Hikvision) & 96.0\% & 91.1\%  & \cmark & -& -\\
   $3^{rd}$ (USTC) & 93.0\% & 95.6\% & \cmark & - & -\\
   $4^{th}$ (Ours w/o PDNC \& CMMT) & 93.0\% & 75.4\%   & \xmark & \xmark & - \\
   $5^{th}$  & 82.0\% & 68.7\%   & - & - &-\\
   \hdashline
   Official Baseline 1: Zhang \etal~\cite{zhang2022voxeltrack}  (TPAMI 22) & 76.8\% & 50.8\%   & - & \xmark &-\\  
   Official Baseline 2: Han \etal~\cite{han2021mmptrack} (ArXiv 21) & 94.6\% & 74.1\%   & \cmark & \xmark &-\\

   \midrule
   \multicolumn{6}{l}{After challenge deadline submissions} \\

   \textbf{New  $3^{rd}$ (Ours w/ PDNC \& CMMT)} & 95.0\% & 84.3\%  & \xmark & \xmark & 67 FPS\\
   \bottomrule
  \end{tabular}
  \end{center}
\end{table*}

\begin{table}[!h]
	\centering
	\setlength{\tabcolsep}{2pt}
	\footnotesize
	\caption{\textbf{Ablation studies for our PNDC on the \textit{WILDTRACK} dataset~\cite{chavdarova2018wildtrack} (w/o appearance)}. MOTA is used as the dominant evaluation metric. The definitions of MOTA and IDF1 are provided in \textit{refs.}~\cite{MOTChallenge2015,milan2016mot16}. The data rendered in \textbf{Bold} and \underline{Underlined} indicate the best and second-best results respectively. }
	\label{tab:multi-wildtrack_ablation}
	\begin{tabular}{lccc}
  \toprule
	Method  & IDF1 $\uparrow$ & MOTA  $\uparrow$ & IDs $\downarrow$ \\ 
	\midrule
	\makecell[l]{Our framework w/ \\complete-linkage clustering} & 87.3\%  & 73.6\%  & 24 \\  \midrule
	Our framework w/ PDNC  & \textbf{94.7\%}  & \textbf{94.0\%} & \textbf{11}  \\ 
	\bottomrule
	\end{tabular}
  \end{table}

  \begin{figure*}[!h]
    \centering
    \includegraphics[width=\textwidth]{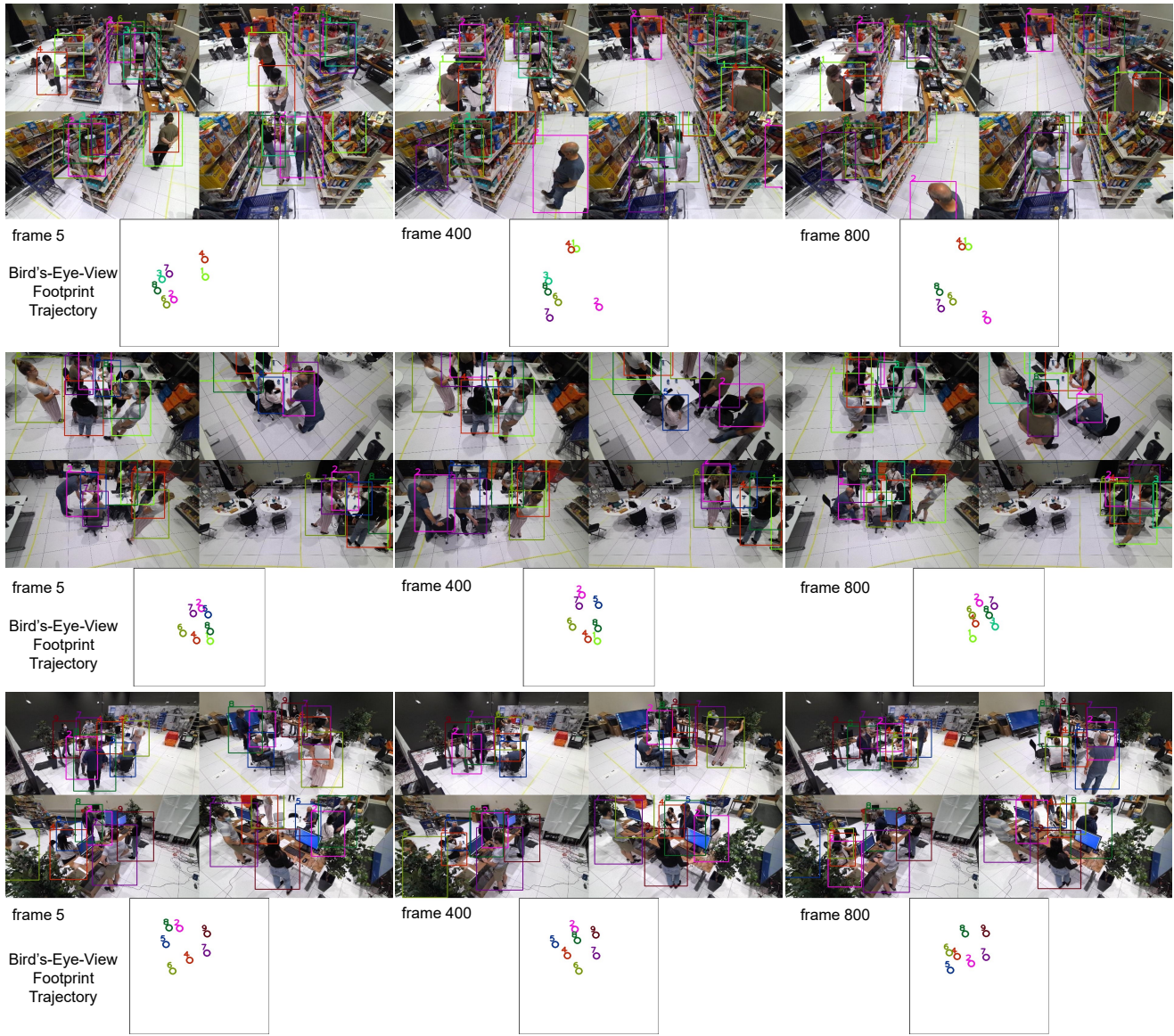}
    \caption{\textbf{Qualitative results on the test set of \textit{MMPTRACK} dataset}~\cite{han2021mmptrack}. 2D bounding boxes are applied to locate persons in each camera view and circles are utilized to indicate the 3D locations of persons in the bird’s-eye view coordinate. Each person is denoted by a unique color based on the tracking results.}\label{fig:results_MMPTRACK}
  \end{figure*}

  \begin{figure*}[!h]
    \centering
    \includegraphics[width=\textwidth]{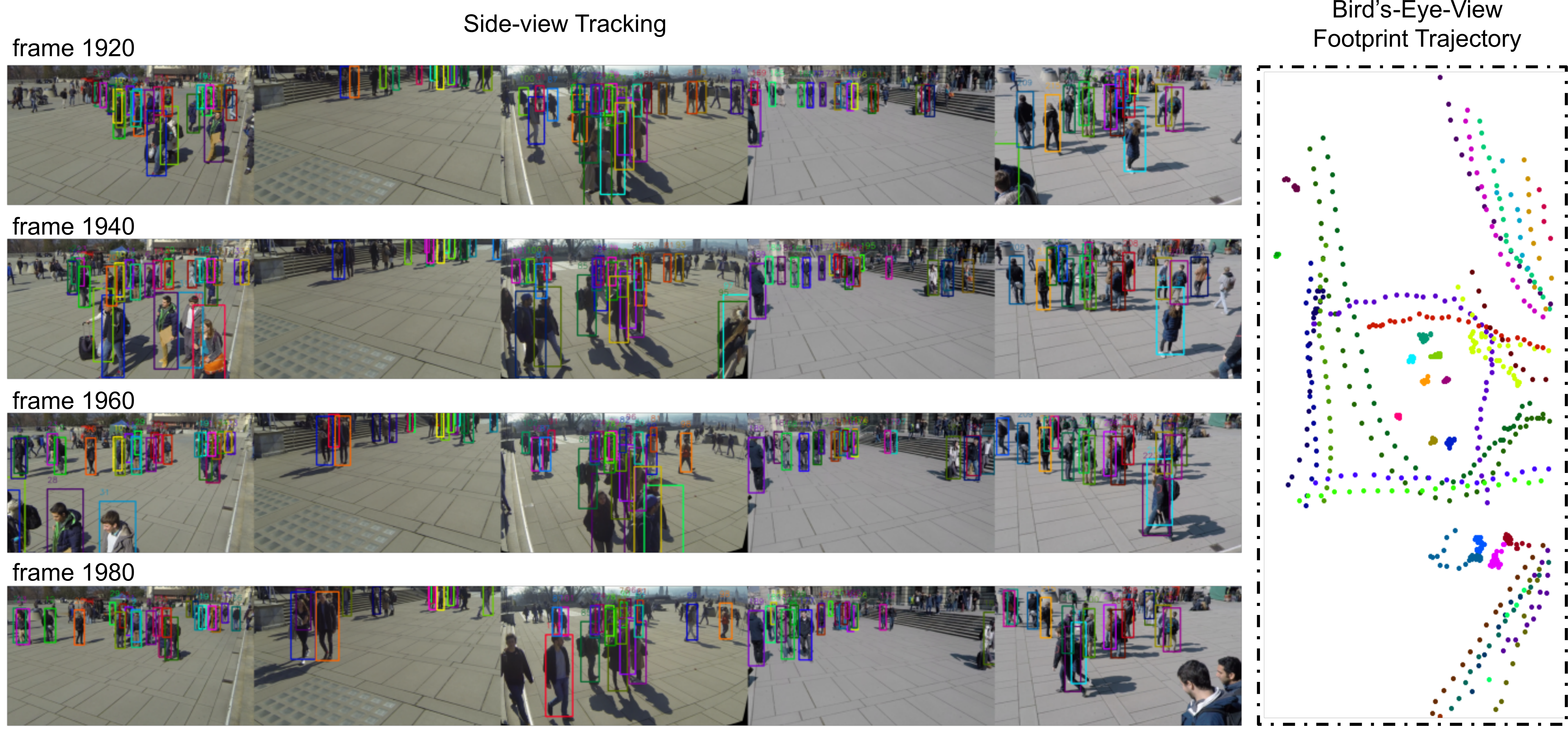}
    \caption{\textbf{Qualitative results on the test set of \textit{WILDTRACK} dataset}~\cite{chavdarova2018wildtrack}. 2D bounding boxes are applied to locate persons in each camera view and circles are utilized to indicate the 3D locations of persons in the bird’s-eye view coordinate. Each person is denoted by a unique color based on the tracking results.}\label{fig:results_WILDTRACK}
  \end{figure*}

\begin{table*}[h!]
  \begin{center}
  \caption{\textbf{Comparison with the state-of-the-art methods on the Campus and the Shelf datasets~\cite{fleuret2007multicamera,BelagiannisAASN16}.} The evaluation metrics are PCP~\cite{fleuret2007multicamera,BelagiannisAASN16} and MOTA~\cite{milan2016mot16}. The higher the PCP and MOTA scores are, the better the performance. The data rendered in \textbf{Bold} indicates the best result.}\label{tab:campus_shelf}
  \begin{minipage}{\linewidth}
  \centering
  \begin{tabular}{rccccc}
  \toprule
  Campus & Actor 1 & Actor 2 &Actor 3 &Average PCP &MOTA  \\ \midrule
  Belagiannis \etal ~\cite{Belagiannis2014} (CVPR 14) & 82.0\% & 72.4\% & 73.7\% & 75.8\%  &-\\ 
  Belagiannis \etal ~\cite{BelagiannisWSFI14} (ECCVW 14) & 83.0\% & 73.0\% & 78.0\% & 78.0\% &- \\ 
  Belagiannis \etal ~\cite{fleuret2007multicamera,BelagiannisAASN16} (TPAMI 16) & 93.5\% & 75.7\% & 84.4\% & 84.5\% &- \\ 
  Ershadi-Nasab \etal ~\cite{Ershadi-NasabNK18} (MTA 18) & 94.2\% & 92.9\% & 84.6\% & 90.6\%  &-\\ 
  Ye \etal ~\cite{ye2022faster} (ECCV 22) & 96.5\% &94.1\% &97.9\% &96.2\%  &-\\ 
  Dong \etal ~\cite{dong2019fast} (CVPR 19) & 97.6\% & 93.3\% & 98.0\% & 96.3\% &-\\ 
  Zhang \etal~\cite{tu2020voxelpose} (ECCV 20) & 97.6\% &  93.8\%  & \textbf{98.8\%} & 96.7\% &-   \\ 
  Zhang \etal~\cite{zhang2022voxeltrack} (TPAMI 22) & 98.1\% &  93.7\%  & 98.3\% & 96.7\% & 89.3\%   \\ 
  \textbf{Ours w/o appearance}  & \textbf{98.2\%} &  \textbf{94.6\%}  & 98.2\% &  \textbf{97.0\%} & 89.4\%  \\ 
  \textbf{Ours w/ appearance}  & \textbf{98.2\%} &  \textbf{94.6\%}  & 98.2\% &  \textbf{97.0\%} & \textbf{90.8\%}  \\ 
  \bottomrule
  \end{tabular}
\end{minipage} 

\begin{minipage}{\linewidth}
  \centering
  \begin{tabular}{rccccc}
  \toprule
  Shelf & Actor 1 & Actor 2 &Actor 3 &Average PCP &MOTA  \\ \midrule
  Belagiannis \etal ~\cite{Belagiannis2014} (CVPR 14) & 66.1\% &65.0\% &83.2\% &71.4\% &-  \\ 
  Belagiannis \etal ~\cite{BelagiannisWSFI14} (ECCVW 14) & 75.0\% &67.0\% &86.0\% &76.0\% &- \\ 
  Belagiannis \etal ~\cite{fleuret2007multicamera,BelagiannisAASN16} (TPAMI 16) & 75.3\% &69.7\% &87.6\% &77.5\% &- \\ 
  Ershadi-Nasab \etal ~\cite{Ershadi-NasabNK18} (MTA 18) & 93.3\% &75.9\% &94.8\% &88.0\% &- \\ 
  Dong \etal ~\cite{dong2019fast} (CVPR 19) & 98.8\% &94.1\% &97.8\% &96.9\% &-\\ 
  Zhang \etal~\cite{tu2020voxelpose} (ECCV 20) & 99.3\% &94.1\% &97.6\% &97.0\% &- \\ 
  Zhang \etal~\cite{zhang2022voxeltrack} (TPAMI 22) & 98.6\% &  94.9\%  & \textbf{97.7\%} & 97.1\% & 94.4\% \\
  Zhang \etal~\cite{zhang20204d} (CVPR 20) & 99.0\% &\textbf{96.2\%} &97.6\% &97.6\% &- \\
  Ye \etal ~\cite{ye2022faster} (ECCV 22) & 99.4\% &96.0\% &97.5\% &97.6\%  &-\\ 
  \textbf{Ours w/o appearance} & \textbf{99.5\%} &  96.0\%  & \textbf{97.7\%} &  \textbf{97.7\%} & \textbf{94.6\%}  \\   
  \textbf{Ours w/ appearance} & \textbf{99.5\%} &  96.0\%  & \textbf{97.7\%} &  \textbf{97.7\%} & \textbf{94.6\%}  \\
  \bottomrule
  \end{tabular}
\end{minipage}
\end{center}
\end{table*}

\begin{table}[h!]
  \setlength{\tabcolsep}{.8pt}
  %\footnotesize
  \caption{\textbf{Ablation studies for our CMMT on the \textit{Campus} and 
  \textit{Shelf} datasets~\cite{fleuret2007multicamera,BelagiannisAASN16} .} The metric is PCP~\cite{fleuret2007multicamera,BelagiannisAASN16}. The larger the PCP score is, the better the performance. The data rendered in \textbf{Bold} indicates the best results. Using the same multi-view associated 2D tracklets, we demonstrated that our CMMT is better than using the combination of Triangulation and RANSAC.}\label{tab:ablation_campus_shelf}
 
  \begin{minipage}{\linewidth}
  \centering
  \begin{tabular}{lcccc}
  \midrule
  Campus & Actor 1 & Actor 2 &Actor 3 &Average    \\ \midrule
  w/ Triangulation & 96.1\%  & 93.0\% & 90.7\% & 93.3\% \\ 
  w/ Triangulation+RANSAC & 97.6\% & 93.0\% & 95.6\% & 95.4\% \\ 
  w/ CMMT (our proposal) & \textbf{98.2\%} &  \textbf{94.6\%}  & \textbf{98.2\%} &  \textbf{97.0\%}   \\ 
  \bottomrule
  \end{tabular}
  \end{minipage}

\begin{minipage}{\linewidth}
  \centering
  \begin{tabular}{lcccc}
  \midrule
  Shelf & Actor 1 & Actor 2 &Actor 3 &Average    \\ \midrule
  w/ Triangulation & 96.6\% & 92.0\%  & \textbf{97.7\%} & 95.4\% \\ 
  w/ Triangulation+RANSAC & 98.5\% & 93.8& \textbf{97.7\%} & 96.6\% \\ 
  w/ CMMT (our proposal)   & \textbf{99.5\%} &  \textbf{96.0\%}  & \textbf{97.7\%} &  \textbf{97.7\%}   \\ 
  \bottomrule
  \end{tabular}
\end{minipage}
\end{table}

\subsection{Experiment Details}

To build a unified 3D MM-Tracking framework that can perform robust 3D pose and footprint tracking, we explore the following aspects of our framework.

\textbf{1. Can our framework generate satisfactory 3D footprints with cross-view 2D bounding boxes?} 

Given that the generation of 2D bounding boxes (\eg, YOLOX~\cite{yolox2021}) could be faster and easier than generating 2D poses (\eg, PoseMachine~\cite{cao2017realtime}) for multiple persons, in some applications, the use of 2D bounding boxes to generate 3D footprints is highly efficient. We investigated the 3D footprint tracking with the \textit{WILDTRACK}~\cite{chavdarova2018wildtrack} and \textit{MMPTRACK}~\cite{han2021mmptrack} datasets.

For the \textit{WILDTRACK}~\cite{chavdarova2018wildtrack} dataset, we show the quantitative and qualitative results in Table~\ref{tab:multi-wildtrack} and Figure~\ref{fig:results_WILDTRACK}, respectively. Our experiments empirically demonstrate that our framework can overcome adverse conditions (\eg, occlusions) and obtain promising results in large-scale outdoor scenes. Compared with the state-of-the-art method, even though we obtain the second-best results in terms of the IDF1 and MOTA scores, our framework offers online processing on realistically sized problems. Besides, we notice that using the appearance features can improve the performance of data association, as the IDF1 score is increased from $94.7\%$ to $97.5\%$, and IDs score is decreased from $11$ to $5$.

For the \textit{MMPTRACK} dataset, the quantitative results are tabulated in Table~\ref{tab:MMPTRACK}, and the qualitative results are illustrated in Figure~\ref{fig:results_MMPTRACK}.
Up to the ICCV MMP-Tracking challenge deadline, our solution ranked the 
fourth place with the MOTA of $93\%$ and IDF1 of $75\%$. At that time, we did not have PDNC and CMMT, but applied conventional Complete-linkage Clustering for cross-view 2D tracklets association and RANSAC for Triangulation outlier rejection. After the challenge deadline, we kept pushing our framework forward by adding PDNC and CMMT. By submitting our new results to the evaluation system, we made an improvement in MOTA ($95\%$) and IDF1 ($84\%$). Overall, we achieved comparable performance to top-ranking methods. For the dominant metric MOTA~(\textit{c.f.},~\cite{MOTChallenge2015,milan2016mot16}), our results reached $95\%$, which is $1\%$ lower than two top-ranking results but should be a promising result by considering the notable crowding and occlusion in the challenging dataset. Besides, the top-ranking solutions had access to the unlabeled test set to boost their detection performance with semi-supervised learning. Our performance could also be improved by utilizing a similar approach, but this is out of the scope of building a unified framework.

For the metric IDF1~(\textit{c.f.},~\cite{MOTChallenge2015,milan2016mot16}), our performance reached a score of $84\%$ and had a margin to match the top-ranking results. However, 3D MM-Tracking is conducted with a complicated system, and a significant number of factors could be varied for the trade-off of the training cost, model generalization, tracking performance, and speed. Unlike other competitors, we did not utilize the appearance feature in our challenge solution. \textit{The challenging dataset only contains a few persons, and their appearances are easily distinguished by the Re-ID model (see Figure~\ref{fig:results_MMPTRACK}). Thus, it is reasonable for competitors, who utilized the appearance cue, to reduce ID switches and obtain better IDF1 scores than us}. Moreover, despite the good performance, the cost of training the Re-ID model is notable and the generalization of the Re-ID model should be considered. On account of the potential domain gaps, the Re-ID model may need to tune its weights to fit the target data set~\cite{sha2020progressive, yang2020remots, yang2021remot}; otherwise, directly applying the Re-ID model may lead to unnecessary degradation of the tracking performance. 
Without using the Re-ID model, we can diminish the additional training cost and speed up the inference process, but we could also sacrifice the performance in terms of the IDF1.

\textbf{2. Does PNDC improve the cross-camera data association?} 

We argued that it is challenging to apply conventional clustering methods to associate cross-view 2D tracklets, because the distance between 2D cross-view tracklets that do not overlap in time cannot be calculated. To solve this problem, we proposed our PDNC. In Table~\ref{tab:multi-wildtrack_ablation}, we show that using our PDNC achieves better tracking performance than using complete-linkage clustering. Furthermore, in our Appendix, we use examples to illustrate the importance of correctly handling 2D cross-view tracklets that do not overlap in time.

\textbf{3. Can our framework generate satisfactory 3D pose trajectories with cross-view 2D poses?} We selected the \textit{Campus} and \textit{Shelf} datasets~\cite{fleuret2007multicamera,BelagiannisAASN16} to evaluate our framework in 3D multi-person pose tracking. 
We follow the same evaluation protocol as in previous works~\cite{dong2019fast, tu2020voxelpose, zhang2022voxeltrack, zhang20204d} to perform a comparison. The 2D detected poses provided by VoxelPose~\cite{tu2020voxelpose} are employed. The PCP~\cite{fleuret2007multicamera, BelagiannisAASN16} and MOTA~\cite{milan2016mot16} are used together as evaluation metrics. 

\begin{figure*}[!h]
  \centering
  \includegraphics[width=0.8\textwidth]{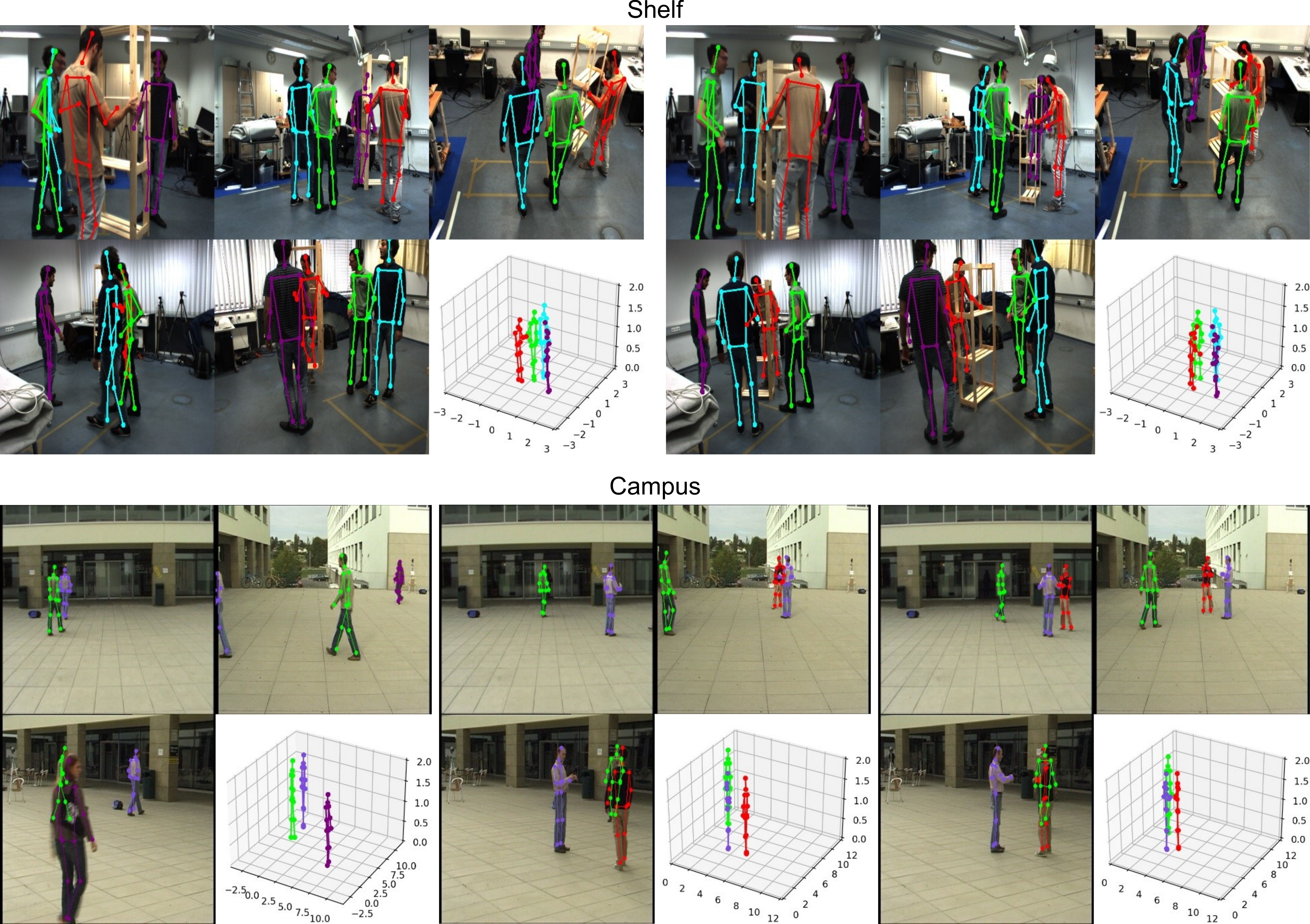}
  \caption{\textbf{Qualitative results on the test set of \textit{Shelf} and \textit{Campus} datasets}~\cite{fleuret2007multicamera,BelagiannisAASN16}. Each person is represented with unique color based on the tracking results.}\label{fig:results_Shelf_Campus}
\end{figure*}

The quantitative evaluation results are shown in Table~\ref{tab:campus_shelf}, and the qualitative evaluation results are illustrated in Figure~\ref{fig:results_Shelf_Campus}. Table~\ref{tab:campus_shelf} shows that our solution outperforms others in terms of the Average PCP. We achieved the average PCP score of $97.0\%$ and $97.7\%$ on the \textit{Campus} and \textit{Shelf} datasets respectively, which are the state-of-the-art performance for these two datasets in terms of 3D pose estimation. From the tracking perspective, our method also reaches the best MOTA score among the reported works. Note that, for \textit{Campus} and \textit{Shelf} datasets, using the appearance feature can slightly improve our tracking performance, but does not affect the 3D pose estimation. Because we already achieved a promising tracking performance with our appearance-free framework.

The samples visualized in Figure~\ref{fig:results_Shelf_Campus} indicate that 3D poses are correctly reconstructed even in heavily occluded scenarios. These results reveal that our framework can generate satisfactory 3D pose trajectories with cross-view 2D poses. Meanwhile, we also proved the effectiveness of our Collaborative Multi-frame Multi-view Association and Collaborative Multi-frame Multi-view Triangulation.

\textbf{4. Does CMMT provide a robust 3D triangulation?} 

On the \textit{Campus} and \textit{Shelf} datasets, we compared our CMMT to two common approaches: (1) solely applying single-frame triangulation and (2) applying single-frame triangulation with RANSAC. For a fair study, we applied the same multi-view associated 2D tracklets to exclude the effect from the previous cross-view association.

We present the results in Table~\ref{tab:ablation_campus_shelf}. Because of the potential errors of the detection and cross-view association, solely using single-frame triangulation might be affected by outliers and the PCP score of 3D positions is relatively lower than that using outlier rejection with RANSAC or our CMMT. The existing approach of applying RANSAC only considers single-frame information for outlier rejection, whereas our CMMT determines the inlier and outliers by consolidating multi-frame multi-view information. The results imply that our CMMT can achieve the best PCP score for 3D positions, and it can be used as a robust 3D triangulation method.

\begin{table*}[h!]
  \begin{center}
  \caption{\textbf{Ablation studies of our parameters on the \textit{MMPTRACK} dataset~\cite{han2021mmptrack} (w/o appearance).} MOTA is used as the dominant evaluation metric. The definitions of MOTA and IDF1 are provided in \textit{refs.}~\cite{MOTChallenge2015,milan2016mot16}. Using the same 2D detections, we demonstrated that our framework is insensitive to the variation of its dominant hyperparameter. The data rendered in \textbf{Bold} indicates the best results.} \label{tab:ablation_MMPTRACK}
  \begin{tabular}{lccc}
    \toprule
    Setting & MOTA$\uparrow$ & IDF1$\uparrow$  &\makecell[c]{Average Speed $\uparrow$}\\
    \midrule
    \multicolumn{4}{l}{Our default setting (w/o appearance).} \\ 
    $\nu=50,\delta=30,\lambda=0.3$ & \textbf{95\%} & \textbf{84\%}   & 67 FPS\\
    \midrule
    \multicolumn{4}{l}{Adjusting window size $\nu$ and step size $\delta$ of the sliding window .} \\

    $\nu=30,\delta=20,\lambda=0.3$ & 94\% & 83\%    & 60 FPS\\
    $\nu=70,\delta=50,\lambda=0.3$ & 94\% & \textbf{84\%}  & \textbf{72 FPS}\\
    \midrule
    \multicolumn{4}{l}{Adjusting multi-view association threshold $\lambda$, which implies the ratio between epipolar distance to target person scale.} \\
   
    $\nu=50,\delta=30,\lambda=0.1$ & 94\% & 81\%   & 67 FPS\\
    $\nu=50,\delta=30,\lambda=0.6$ & \textbf{95\%} & 83\%   & 67 FPS\\
   \bottomrule
  \end{tabular}
  \end{center}
\end{table*}

\textbf{5. How does the dominant hyperparameter affect the model performance?} 

Since the sliding windows are used to traverse videos in 3D MM tracking, the hyperparameter of the sliding windows could affect the tracking result and speed. Besides, the threshold of associating multi-view 2D tracklets plays a key role in our framework. Therefore, we conducted ablation studies to explore the effect of varying $\nu$ (\ie, the window size of the sliding window), $\delta$ (\ie, the step size of the sliding window), and  $\lambda$ (\ie, the threshold of associating multi-view 2D tracklets). In the ablation studies (see Table~\ref{tab:ablation_MMPTRACK}), we vary the value of the hyperparameters within a reasonable range based on their intuitive physical meanings.

In Table~\ref{tab:ablation_MMPTRACK}, the inference speed is increased by increasing the size and step of the sliding window, and vice versa by decreasing the scale and step of the sliding window. Increasing the size and step of the sliding window can eliminate the redundant computation in our Collaborative Multi-frame Multi-view
Association, but it will increase the delay of our near-lone processing. When the window size is equal to the entire video, our framework becomes an offline method, however, when the window size decreases to one frame, our framework becomes an online method. We made a trade-off for the size and step of the sliding window by selecting $\nu=50$ and $\delta=30$. As we have discussed previously, we determine the value of $\lambda$ by referring to the possible ratio between epipolar distances and a person's projection scale. If the detection is perfect, then the epipolar distance should be zero. However, in real practice, due to the detection bias, we assume that the bias should be within $0.3$ of the person's projection scale. We have shown that setting $\lambda=0.1$ and $\lambda=0.6$ may not significantly affect the association performance.

\textbf{6. What is the speed of this framework?}

Although it is common to report an average speed of 3D MM-Tracking for processing an entire dataset, we suggest that it may not provide sufficient knowledge to apply the 3D MM-Tracking system to real applications, because the number of cameras and persons could differ based on the application scenarios. Therefore, we attempted to systematically analyze the speed of our tracking system (w/o appearance) by setting different numbers of cameras and persons. 

\begin{figure}[!h]
  \centering
  \includegraphics[width=\linewidth]{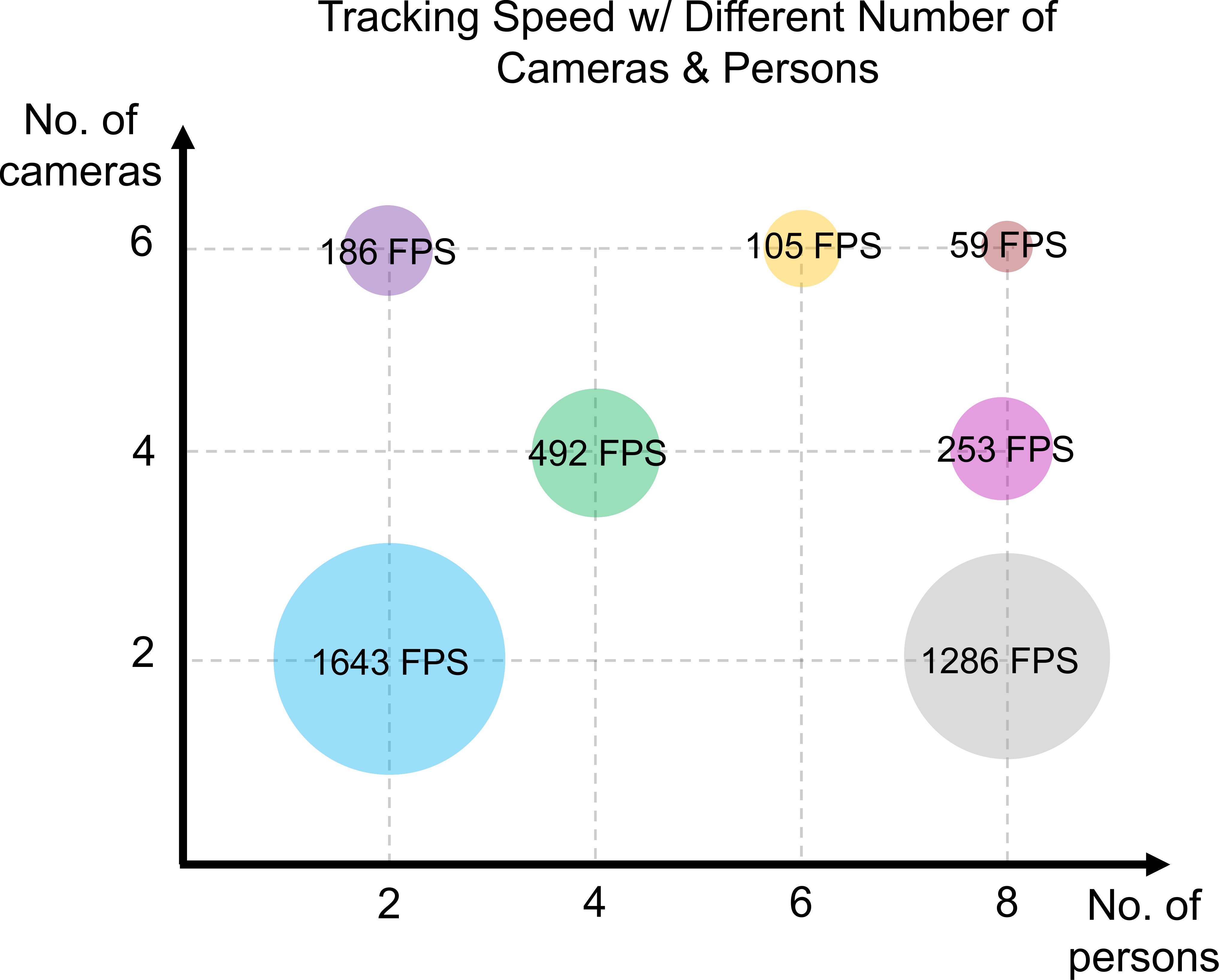}
  \caption{\textbf{The speed of our framework (w/o appearance features).}. The speed of our framework decreases when more cameras and persons are included. }\label{fig:speed}
\end{figure}

The results are shown in Figure~\ref{fig:speed}.
When tracking two persons with two cameras, our framework can achieve a superfast speed of $1,643$ FPS. The speed of our framework decreased to $186$ FPS when tracking two persons with six cameras and $1,286$ FPS when tracking eight persons with two cameras, indicating that the speed of our framework drops when more cameras and persons are included. Nonetheless, since most of the data take four cameras for 3D MM-Tracking, our framework still can be applied to real-time processing.

\begin{figure*}[!h]
  \centering
  \includegraphics[width=0.75\textwidth]{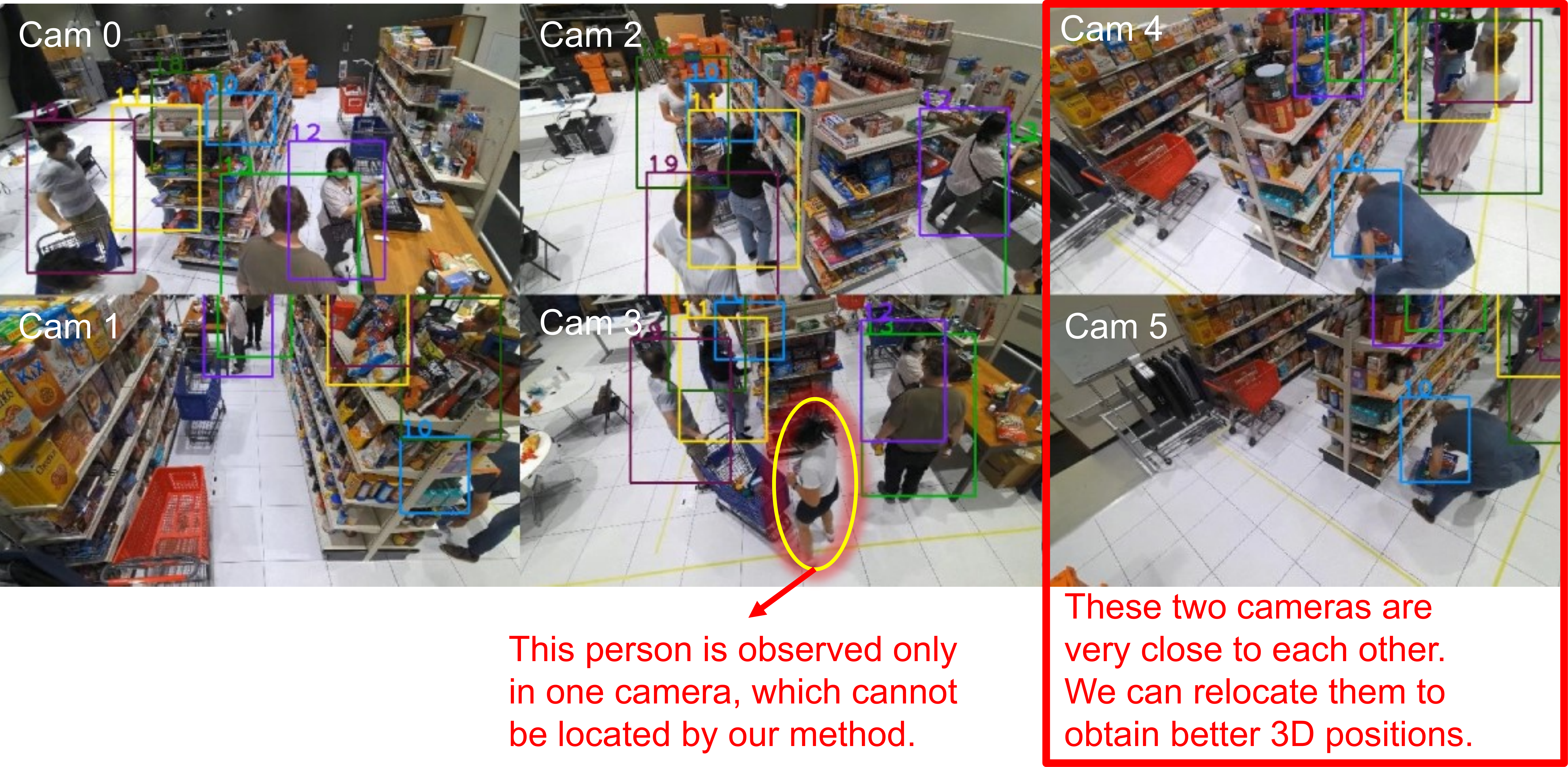}
  \caption{\textbf{Failure case analysis on \textit{MMPTRACK} dataset}~\cite{han2021mmptrack}. Our method cannot handle two corner cases. (1) When only one camera captures the target, the corresponding 3D position cannot be calculated since multi-view stereo pair is unavailable. To solve this issue, we should add more cameras to observe those areas. (2) When two cameras are very close to each other, the triangulation bias will be significantly increased. To ameliorate this issue, we can relocate the cameras to increase the length and angle of the baseline.}\label{fig:failure_case_analysis_MMPTRACK}
\end{figure*}

\textbf{7. What are the limitations of this framework?} 

First, to achieve real-time processing, our recommended framework does not utilize the appearance feature. Although we can decrease the additional training cost in the Re-ID model and increase the inference speed of the whole framework, the ID switches could be relatively higher than using the appearance feature. Nonetheless, as shown in Tables~\ref{tab:multi-wildtrack} and \ref{tab:campus_shelf}, the appearance feature can be easily added to our framework to improve the tracking performance by sacrificing the inference speed. We suggest considering these trade-offs and using the Re-ID model based on the needs.

Second, we selected a typical sample, as shown in Figure~\ref{fig:failure_case_analysis_MMPTRACK}, to explain another limitation of our framework. Different from previous works\cite{sternig2011multi,wen2017multi,he2020multi}, in which the 3D position on the ground plane can be directly generated from a monocular 2D position, our framework brings the multi-view stereo pair of 2D positions to generate the corresponding 3D position. When multi-view stereo pairs cannot be formed, our framework fails to produce 3D positions. To mitigate the problem, we come up with requirements for camera settings. First, we prefer that the overlapping areas of the cameras can cover the entire observation space. Hence, regardless of where a person is, he/she should be simultaneously captured by at least two cameras. Second, we would like to set up sufficient lengths and angles of the baseline between cameras. For multi-view stereo triangulation, it is challenging to obtain accurate 3D positions from multi-view stereo pairs if their corresponding cameras are very close to each other~\cite{hartley2003multiple}. If these requirements are satisfied, the performance of our framework could be further improved.

\section{Conclusion}\label{sec:conclusion}

In this work, we proposed a Unified 3D Multi-view Multi-person Tracking framework, which could be useful in a wide range of downstream applications, such as public spaces, retail stores, and office buildings. Our framework has clear benefits: flexibility, simplicity, and efficiency. First, our framework offers the maximum flexibility to utilize different monocular features. Without additional modifications, our framework can adopt monocular 2D bounding boxes and 2D poses as inputs to produce 3D trajectories for multiple persons. It is also applicable to an arbitrary 3D position that is either in nonplanar or planar environments. Second, although the structure of our framework looks complicated, its hyperparameter is more intuitive than other related works, thereby ensuring simpler utilization. And finally, we verified its effectiveness by accomplishing state-of-the-art performance on the \textit{Campus} and \textit{Shelf} datasets for 3D pose tracking, and comparable results on the \textit{WILDTRACK} and \textit{MMPTRACK} datasets for 3D footprint tracking. Since we have not yet applied the existing useful tricks and modules in our framework, it could serve as a prototype and could be extended to a more powerful 3D multi-view multi-person tracking system in future work.

\appendix
\begin{figure*}[!h]
  \centering
  \includegraphics[width=\textwidth]{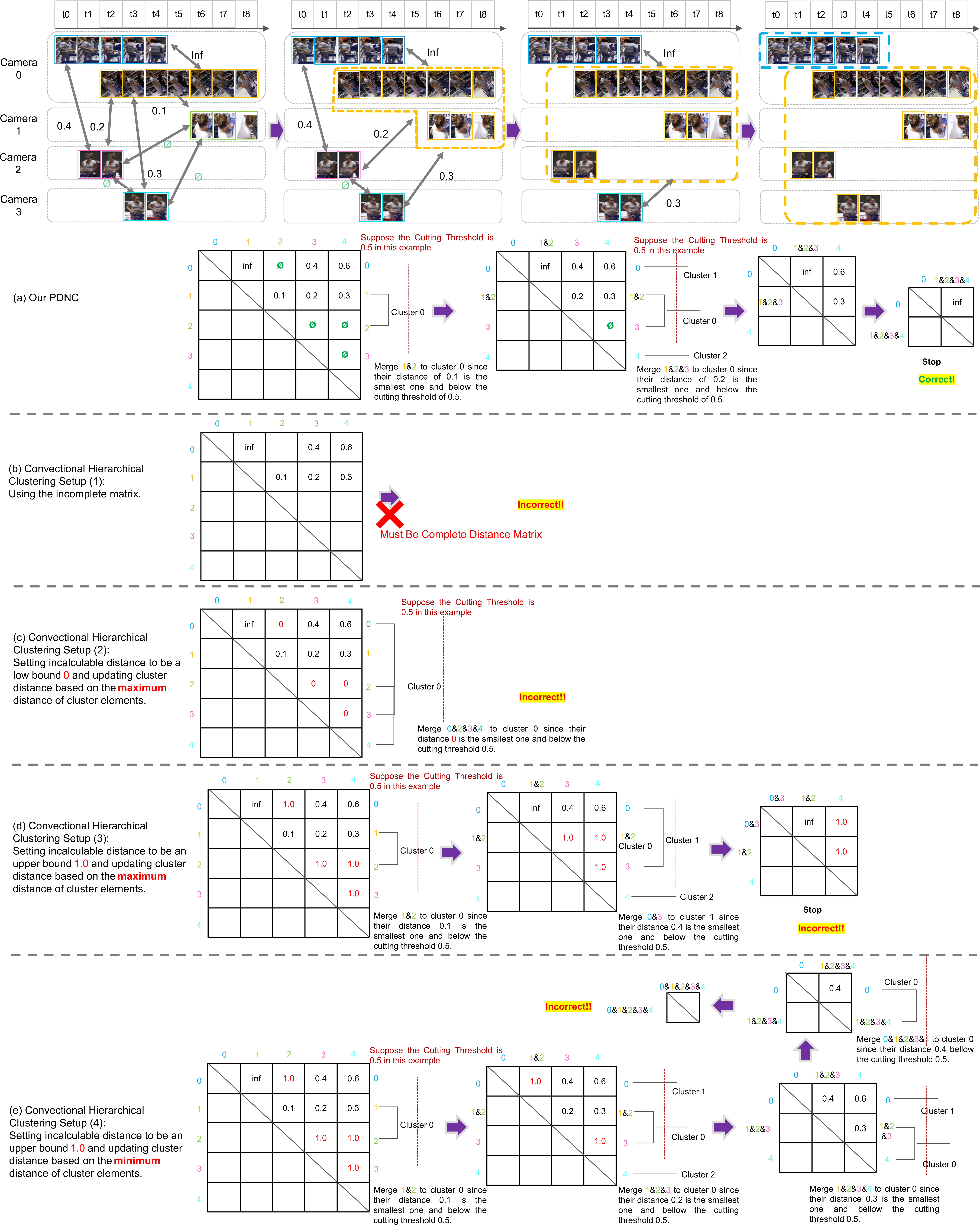}
  \caption{\textbf{Comparison of the clustering details between our PDNC and conventional hierarchical clustering methods.} In this example, we use four camera videos that captured two people, and the goal is to cluster the 2D tracklets based on their epipolar distance. We compare five types of possible strategies, from (a) to (e), in terms of non-parametric clustering. Only our PDNC generates the correct clustering results by properly handling the incalculable distance.}\label{fig:compare_clustering}
\end{figure*}

\subsection*{Appendix}
In this appendix, we illustrate the clustering details of our core proposal---PDNC---on an example of the \textit{MMPTRACK} dataset. In addition, we explain why it is challenging to apply conventional hierarchical clustering methods in multi-view data association. As shown in Figure~\ref{fig:compare_clustering}, supposing four cameras were used to capture two people, the goal is to cluster the 2D tracklets of four camera views based on their epipolar distance. Since the number of persons is unknown for the tracking algorithm, non-parametric clustering should be applied to automatically learn the number of clusters (\ie, the number of persons). We compare five types of possible strategies, from (a) to (e), to explain the necessity and rationality of proposing our model. Note that, for visualization purposes, we double the epipolar distance values and set the cutting threshold $\lambda=0.5$ in Figure~\ref{fig:compare_clustering}.

\noindent \textbf{Strategy (a)} is our PDNC. As presented previously, when there is no temporal overlap between a pair of 2D tracklets, we cannot compute their relative distance. To address this problem, we define the incalculable distance as $\emptyset$ and propose a new PDNC to cluster such a heterogeneous distance matrix. 

\noindent \textbf{Strategy (b)} represents leaving the incalculable distance empty. However, it is challenging to apply existing non-parametric clustering methods to such an incomplete distance matrix.

\noindent \textbf{Strategy (c)} attempts to set the incalculable distance as a low bound with the value of $0$. In the first step of clustering, 2D tracklets $0$, $2$, $3$, and $4$, are incorrectly grouped in the same cluster, even though they contain two persons.

\noindent \textbf{Strategy (d)} is opposite to strategy (c), instead of assigning the incalculable distance as a low bound value, it employs an upper bound value. Supposing all epipolar distances are smaller than $1.0$, then we can use $1.0$ as the upper bound and assign it as the incalculable distance. For cluster distance updating, strategy (d) takes the maximum distance of the cluster element. Nonetheless, 2D tracklets $0$ and $3$ are clustered into the same cluster, which is incorrect.

\noindent \textbf{Strategy (e)} is similar to strategy (d), which also sets the incalculable distance as an upper bound value of $1.0$. While strategy (d) updates the cluster distance based on the maximum distance of the cluster element, strategy (e) selects the minimum distance of cluster element for cluster distance updating. As a result, all 2D tracklets are incorrectly clustered into one cluster.

The results show that our PDNC can generate a correct result by properly handling the incalculable distance.

%\subsection*{Author contributions}
%Fan Yang conceived and designed the study. All authors analyzed the data, discussed the method, and participated in writing the manuscript.

\subsection*{Declaration of competing interest}
The authors have no competing interests to declare that are relevant to the content of this article.

% for bibtex

%\bibliographystyle{CVMbib}
\bibliographystyle{unsrt}
\bibliography{refs}

\end{document}